\documentclass[lettersize,journal]{IEEEtran}
\usepackage{amsmath,amsfonts}
\usepackage{algorithmic}
\usepackage{algorithm}
\usepackage{array}
\usepackage[caption=false,font=normalsize,labelfont=sf,textfont=sf]{subfig}
\usepackage{textcomp}
\usepackage{stfloats}
\usepackage{url}
\usepackage{verbatim}
\usepackage{graphicx}
\usepackage{cite}
\usepackage{multicol,multirow}
\begin{document}

\title{SUNet: Scale-aware Unified Network for\\ Panoptic Segmentation}

\author{Weihao Yan,
  Yeqiang Qian,~\IEEEmembership{Member,~IEEE,} 
  Chunxiang Wang,~\IEEEmembership{Member,~IEEE,}
  and Ming Yang,~\IEEEmembership{Member,~IEEE}
\thanks{This work is supported by the National Natural Science Foundation of China (62103261/62173228). \emph{(Corresponding author: Ming Yang.)}}
\thanks{Weihao Yan, Chunxiang Wang and Ming Yang are with the Department of Automation, Shanghai Jiao Tong University, Shanghai, 200240; Key Laboratory of System Control and Information Processing, Ministry of Education of China, Shanghai, 200240; Shanghai Engineering Research Center of Intelligent Control and Management, Shanghai 200240, China. (email: ywh926934426@sjtu.edu.cn; wangcx@sjtu.edu.cn; mingyang@sjtu.edu.cn)}
\thanks{Yeqiang Qian is with University of Michigan-Shanghai Jiao Tong University Joint Institute, Shanghai Jiao Tong University, Shanghai, 200240, China. (email: qianyeqiang@sjtu.edu.cn)}}

\markboth{Journal of \LaTeX\ Class Files,~Vol.~14, No.~8, August~2021}%
{Yan \MakeLowercase{\textit{et al.}}: SUNet: Scale-aware Unified Network for Panoptic Segmentation}


\maketitle

\begin{abstract}
  Panoptic segmentation combines the advantages of semantic and instance segmentation, 
  which can provide both pixel-level and instance-level environmental perception information for intelligent vehicles.
  However, it is challenged with segmenting objects of various scales, especially on extremely large and small ones.
  In this work, we propose two lightweight modules to mitigate this problem. 
  First, Pixel-relation Block is designed to model global context information for large-scale things, 
  which is based on a query-independent formulation and brings small parameter increments. 
  Then, Convectional Network is constructed to collect extra high-resolution information for small-scale stuff,
  supplying more appropriate semantic features for the downstream segmentation branches. 
  Based on these two modules, we present an end-to-end Scale-aware Unified Network (SUNet), which is more adaptable to multi-scale objects. 
  Extensive experiments on Cityscapes and COCO demonstrate the effectiveness of the proposed methods.
\end{abstract}


\section{Introduction}
\IEEEPARstart{T}{he} capability of understanding the driving environment from visual data is important for intelligent vehicles. 
Semantic and instance segmentation are two classic scene understanding tasks. 
They focus on background stuff and foreground things, respectively, only partially understanding the scene. 
Panoptic segmentation \cite{kirillov2019panoptic} combines advantages of them, 
which requires assigning each pixel of the image with a semantic label and an instance id.
Objects are divided into two categories in panoptic segmentation. 
Countable objects like cars and people are called things, whereas amorphous and uncountable regions like sky and road are called stuff. 
Panoptic segmentation can provide pixel-level semantic information for the entire image
while distinguishing things at the target level, which is beneficial for the safe driving of intelligent vehicles.

\begin{figure}[t]
	\captionsetup[subfloat]{labelformat=empty,font=scriptsize,labelfont=scriptsize}
	\centering
  	\subfloat{\includegraphics[width=0.48\linewidth]{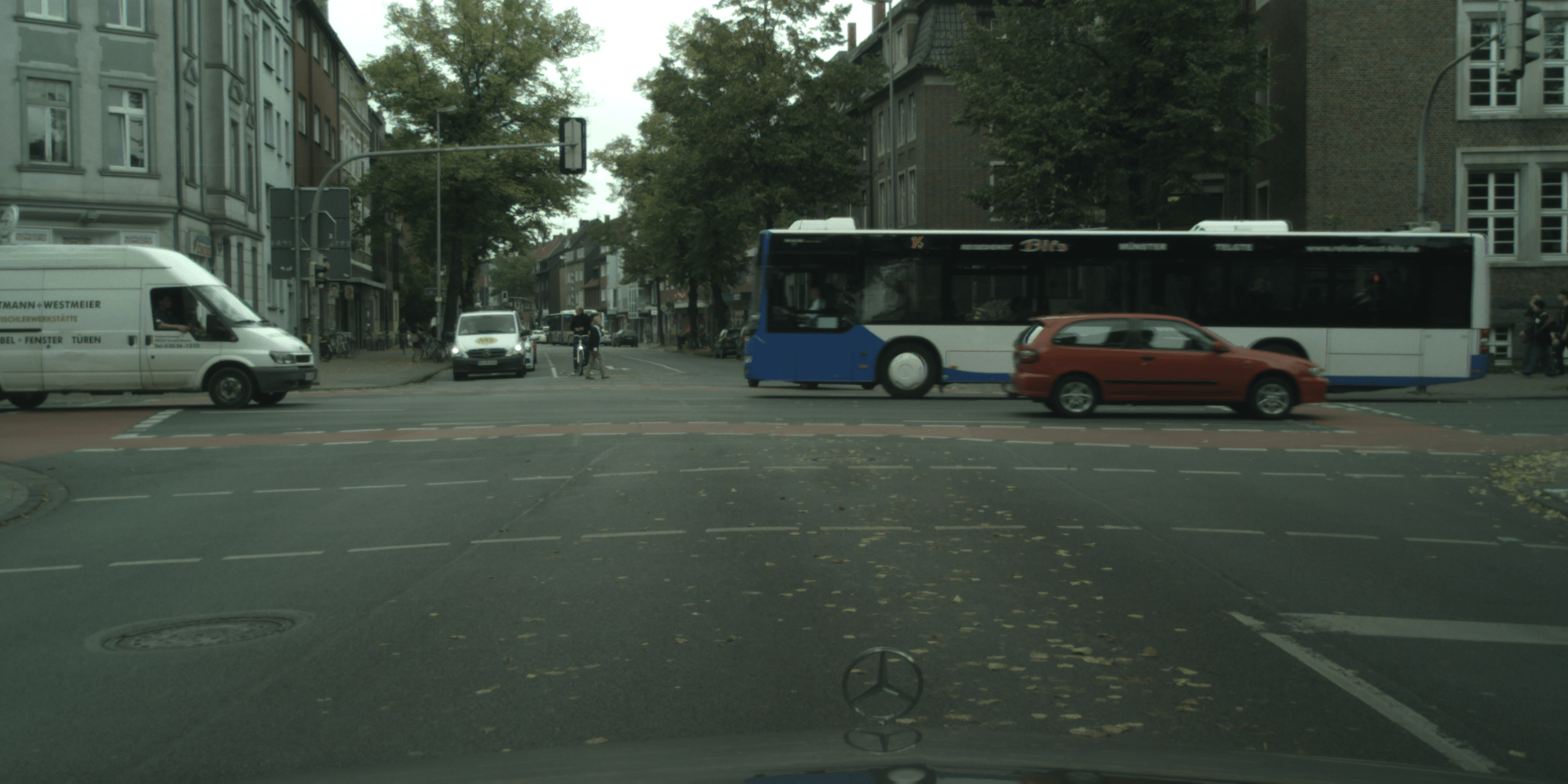}}\hspace{0.02cm}
	\subfloat{\includegraphics[width=0.48\linewidth]{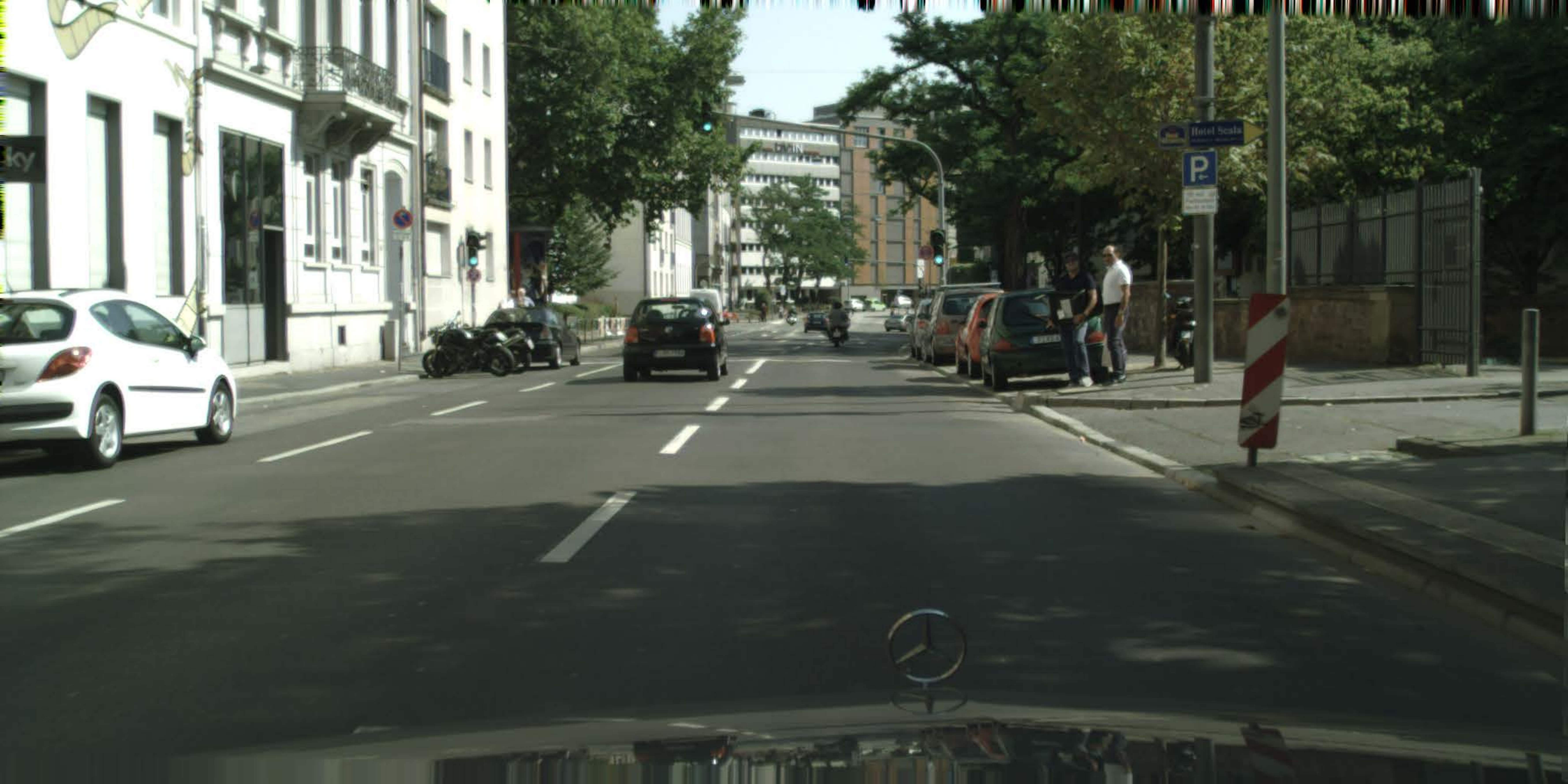}}\vspace{0.1cm}
	\subfloat{\includegraphics[width=0.48\linewidth]{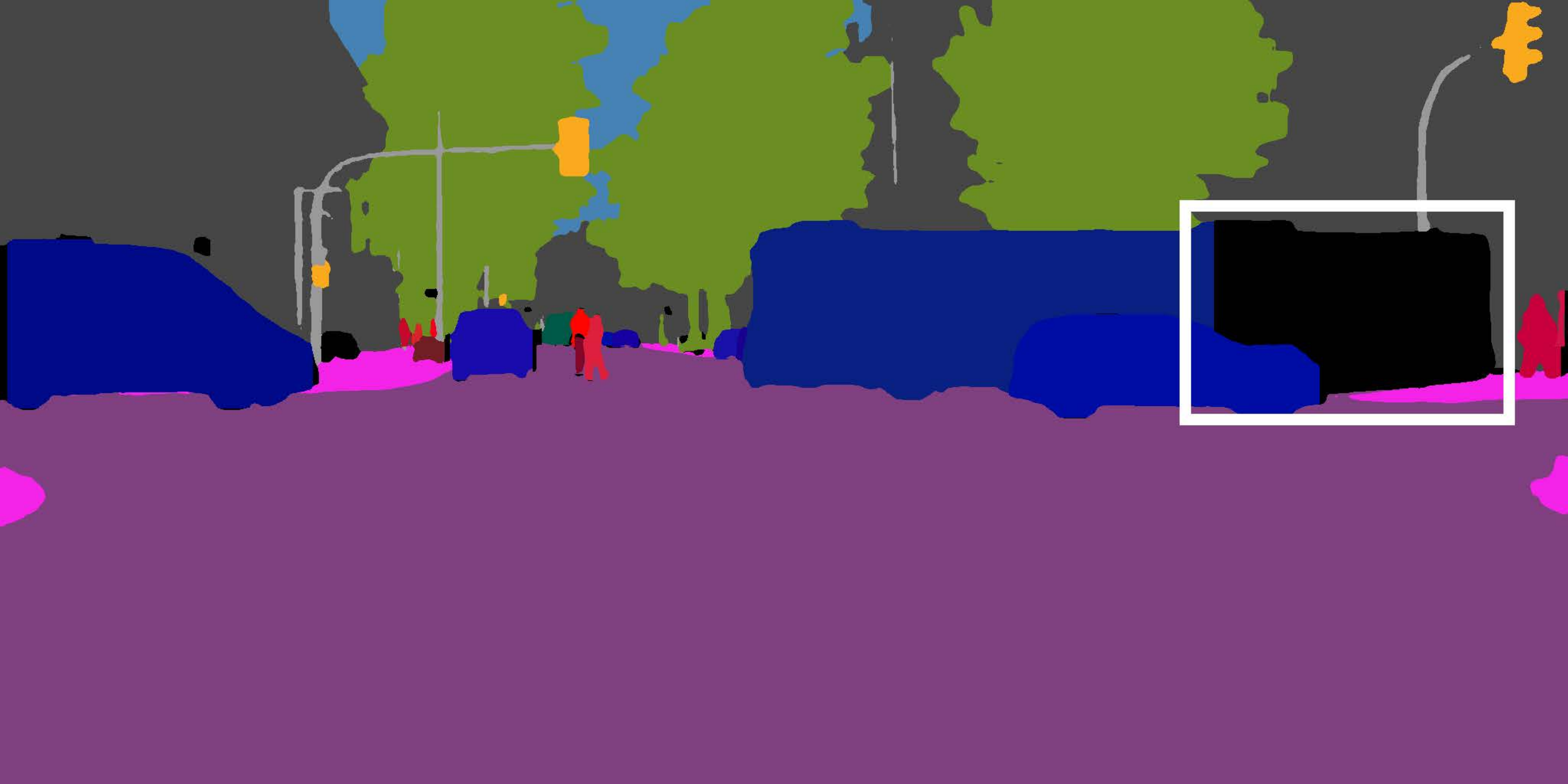}}\hspace{0.02cm}
	\subfloat{\includegraphics[width=0.48\linewidth]{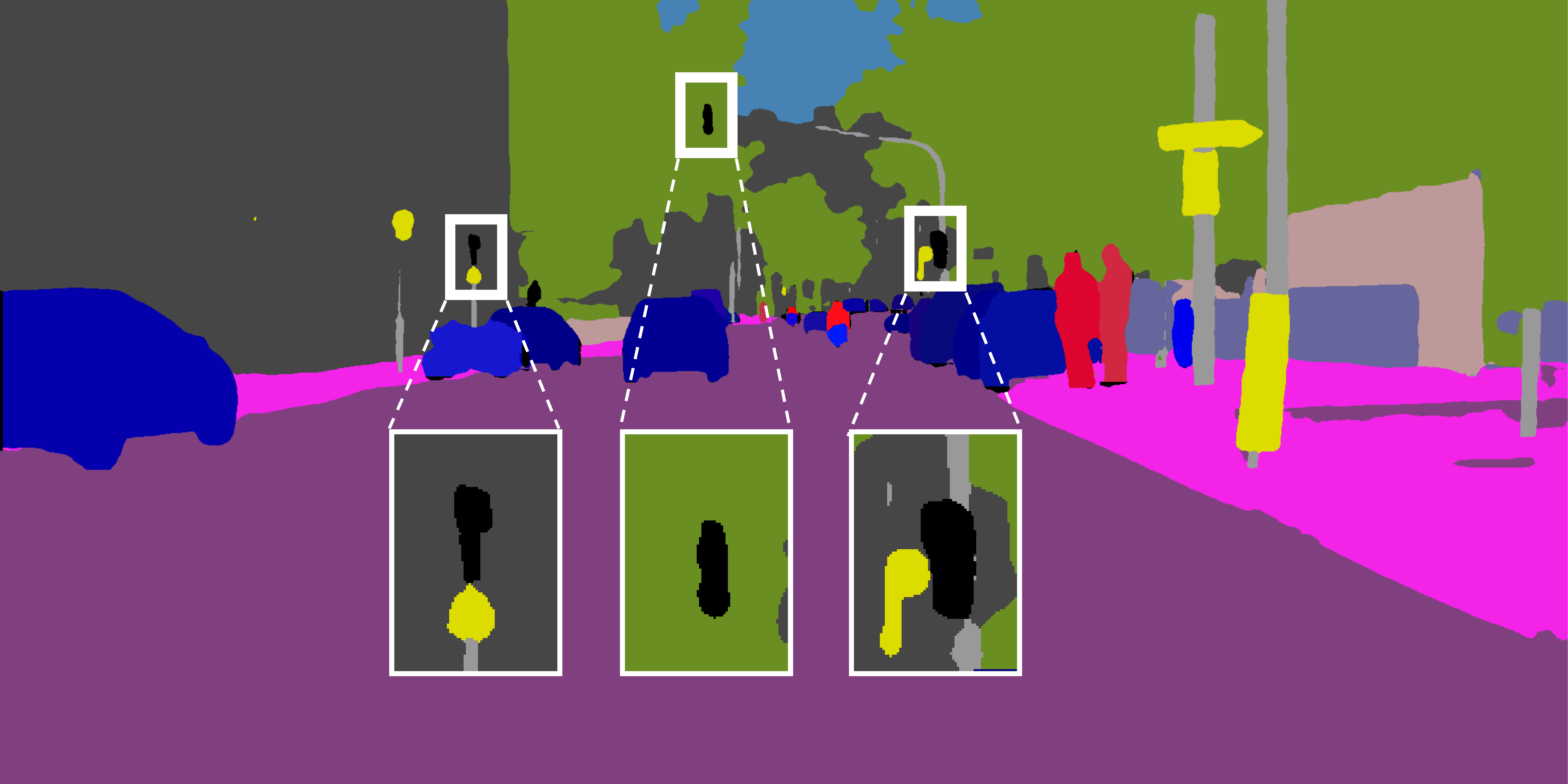}}\vspace{0.1cm}
  	\subfloat[(a)]{\includegraphics[width=0.48\linewidth]{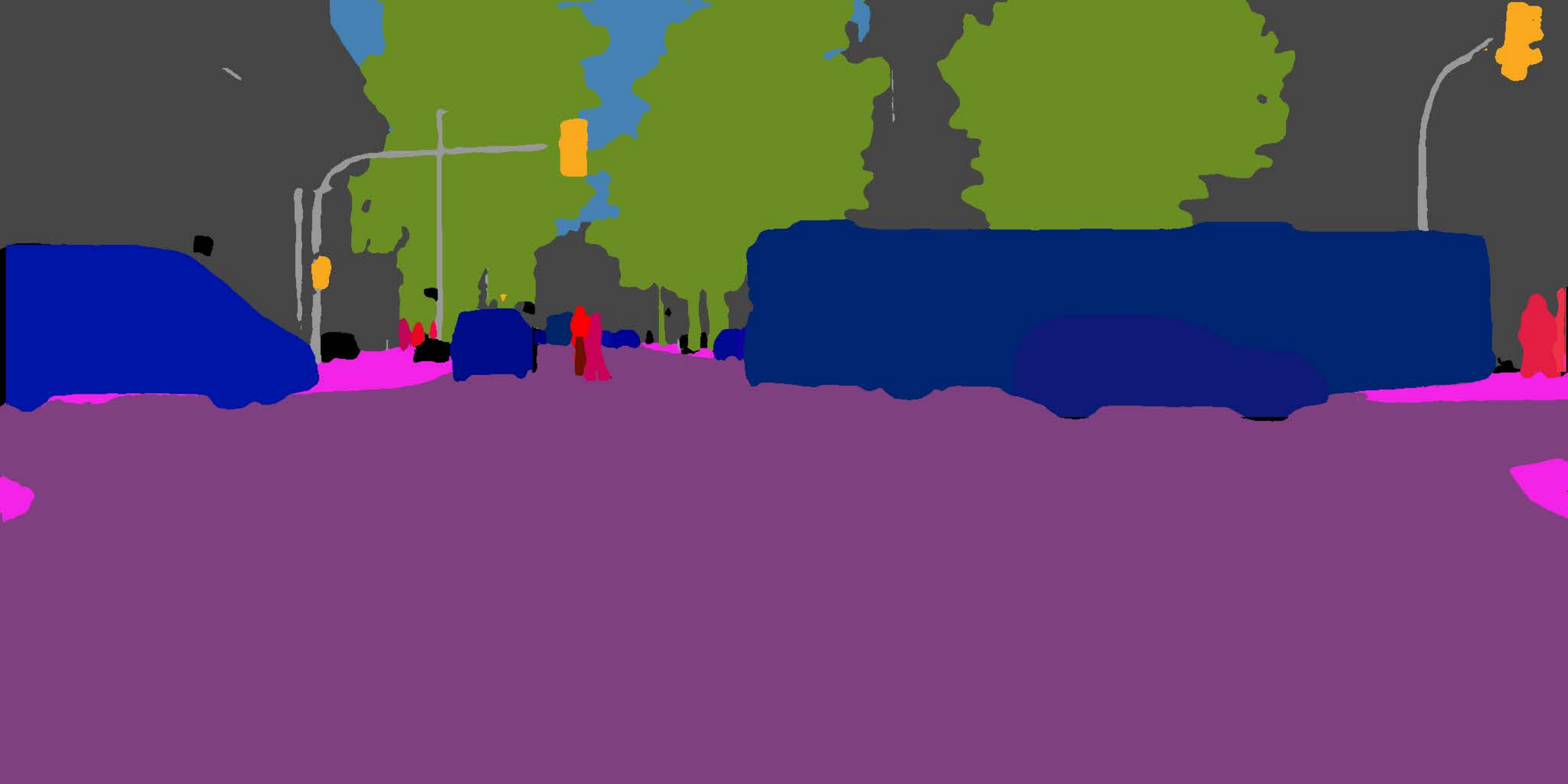}}\hspace{0.02cm}
  	\subfloat[(b)]{\includegraphics[width=0.48\linewidth]{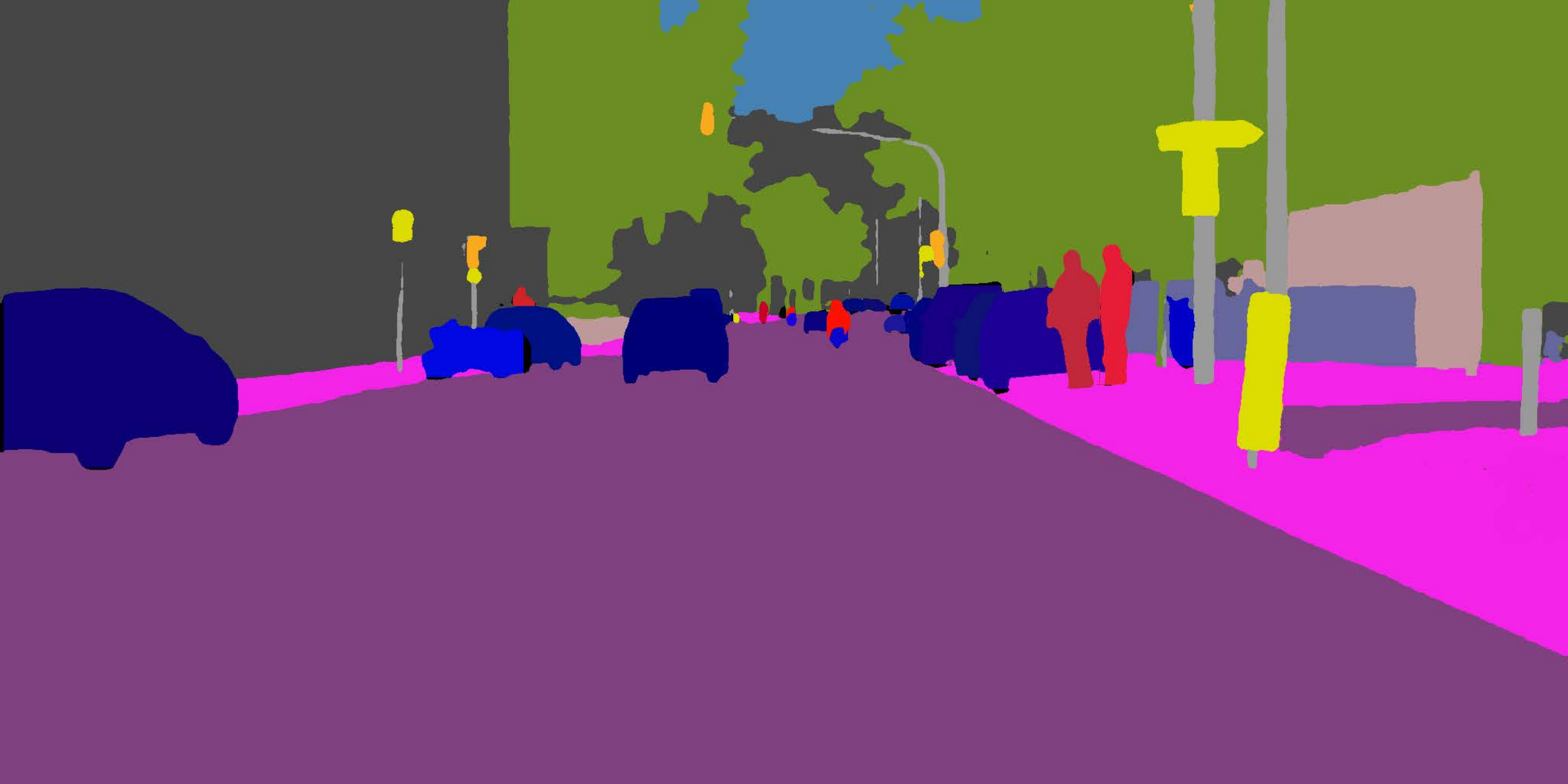}}
	\caption{Visualization of the problems when encountering multi-scale objects in panoptic segmentation. 
	(a) Truncated detection of large-scale things. 
	(b) Poor segmentation on small-scale stuff. 
	From top to down are raw images in Cityscapes, predictions of UPSNet\cite{xiong2019upsnet}, and predictions of SUNet, respectively. 
	Problems are marked by white boxes, and small objects are scaled up onto the road for better viewing.}
	\label{fig:problem_show}
\end{figure}

The first panoptic segmentation method\cite{kirillov2019panoptic} 
resorts to two individual networks for semantic and instance segmentation. 
Current methods usually construct two segmentation branches on a shared backbone to improve the overall efficiency\cite{chen2020banet,li2019attention,wu2020bidirectional,panopticdeeplab,pixelvoting,hou2020realtime,liu2019oanet,xiong2019upsnet,seamless,utips,panopticfpn,efficientps,FastPSNet1}.
However, these works still face the challenge of segmenting multi-scale objects, 
which is even highlighted when encountering extremely large and small objects, as shown in Fig.~\ref{fig:problem_show}.
Some methods\cite{panopticdeeplab,pixelvoting,efficientps,allscales} tend to improve the performance on multi-scale objects by 
utilizing stronger and usually more complicated semantic or instance segmentation models.
Nevertheless, it is not consistent with the original intention of panoptic segmentation jointly considering these
two scene understanding tasks and usually brings a lot of parameters and computation.
In this paper, we transfer the attention to the backbone and 
intend to develop lightweight modules to mitigate this problem.

In deep neural network, 
the receptive fields of the top layer feature maps are restricted\cite{deng2019restricted}. 
For large-scale objects, such as vehicles with large image areas or aspect ratios, their complete feature information cannot be effectively extracted. 
Meanwhile, global context information and long-range dependence\cite{nonlocal}, 
which are essential for the detection of large-scale things\cite{relationnetwork}, 
are not modeled by most existing methods\cite{chen2020banet,li2019attention,wu2020bidirectional,panopticdeeplab,pixelvoting,hou2020realtime,liu2019oanet,xiong2019upsnet,seamless,utips,panopticfpn,efficientps,FastPSNet1} mainly due to a large amount of computation.
Furthermore, the poor segmentation of small-scale stuff should also be noticed.
Though it exists in traditional segmentation tasks, there is another inherent reason in panoptic segmentation.
Current methods\cite{xiong2019upsnet,chen2020banet,li2019attention,wu2020bidirectional,pixelvoting,liu2019oanet,lazarow2020learninginstance,seamless,panopticfpn,utips,FastPSNet1,efficientps,allscales} usually add an extra semantic segmentation branch to instance segmentation model without considering the underlying differences between them: 
instance segmentation tends to utilize object-level features, while semantic segmentation requires refined details. 
However, the Feature Pyramid Network (FPN)\cite{fpn} mainly delivers strong semantic features\cite{efficientps}. 
These two problems are illustrated in Fig. \ref{fig:problem_show}. 
The large-scale bus is not completely detected in Fig. \ref{fig:problem_show}(a), and small traffic signs are not segmented in Fig. \ref{fig:problem_show}(b). 
These problems will cause severe consequences in real applications like autonomous driving. 

In this paper, we propose Pixel-relation Block and Convectional Network to handle these two cases. 
On the one hand, Pixel-relation Block introduces the idea of global context modeling into panoptic segmentation
and has a concise and effective modeling procedure. 
On the other hand, Convectional Network is constructed for small-scale stuff, 
which provides additional high-resolution information for the semantic branch and is more appropriate for sharing by the downstream segmentation branches.  
From quantitative and qualitative experimental results, Pixel-relation Block and Convectional Network are verified to be efficient and effective. 
Moreover, we build Scale-aware Unified Network (SUNet), which is more adaptable to multi-scale objects. 
Experimental results demonstrate that the SUNet has competitive panoptic segmentation performance on the Cityscapes and COCO datasets.

The main contributions of this paper are as follows:

\begin{itemize}
	\item We introduce the idea of global context modeling into panoptic segmentation and design the lightweight Pixel-relation Block,
	which alleviates the truncated detection of large-scale things.
	\item We propose the Convectional Network, which considers the requirements of downstream segmentation tasks
	and supplies more appropriate semantic features.
	\item We build Scale-aware Unified Network (SUNet). 
	It achieves competitive panoptic segmentation performance on Cityscapes and COCO datasets.
\end{itemize}

\section{Related work}
\subsection{Semantic Segmentation}
Semantic segmentation performs segmentation at pixel-level and lacks information at target-level. 
FCN\cite{fcn} uses continuous convolutional layers, firstly achieving pixel-level segmentation. 
It is used as the semantic segmentation branch in \cite{xiong2019upsnet,li2019attention,liu2019oanet}. 
The work in\cite{shadow} considers shadow detection and removal for illumination consistency on the road.
OPP-net\cite{deng2017cnn} is deliberately constructed for the vast variations of objects in the fisheye image. 
The deeplab-like module is designed in\cite{seamless}, and DeeplabV3\cite{deeplabv3} is utilized in\cite{allscales} to improve the panoptic segmentation on stuff categories. 
What's more, GRBNet\cite{gated} proposes a gated-residual block to effectively fuse RGB and depth signals. 
For lightweight and fair comparison, FCN is adopted in our method.

\subsection{Instance Segmentation}
Instance segmentation usually focuses on target-level information and lacks background information. 
Instance segmentation methods can be divided into two categories: one-stage and two-stage. 
In one-stage methods, the candidate regions are usually not proposed. 
They are applied to reduce the prediction time of panoptic segmentation models\cite{panopticdeeplab,pixelvoting}. 
In two-stage methods, the first stage proposes regions of interest. 
The second stage usually has multiple branches, which output the bounding boxes and category labels of targets. 
Among these methods, Mask R-CNN\cite{maskrcnn} is the predominantly used instance segmentation network in panoptic segmentation 
and is also utilized in our network for a fair comparison.

\subsection{Panoptic Segmentation}
Panoptic segmentation is a new potential scene understanding task,
which combines the advantages of semantic and instance segmentation. 
The first panoptic segmentation method\cite{kirillov2019panoptic} uses PSPNet\cite{pspnet} and Mask R-CNN parallelly and a heuristic merging algorithm to fuse the results. 
After that, Panoptic Feature Pyramid Networks\cite{panopticfpn} combines two backbones into one and replaces PSPNet with FCN to reduce parameters and calculations. 
It serves as a strong baseline for the following panoptic segmentation methods.


Stronger semantic or instance segmentation models are utilized in\cite{panopticdeeplab,pixelvoting,efficientps,allscales} to improve panoptic segmentation results. 
DeeplabV3\cite{deeplabv3} is utilized in\cite{allscales} for better semantic segmentation performance. 
Enhanced Mask R-CNN is created in\cite{efficientps} for more accurate instance segmentation. 
However, these methods are more inclined to improve panoptic segmentation results by advancing the performance of semantic or instance segmentation networks. 
It is not in accord with the original intention of panoptic segmentation.

Some methods\cite{liu2019oanet,pixelvoting,lazarow2020learninginstance,xiong2019upsnet} intend to solve the problem of instances occlusion, 
which is due to the differences between instance segmentation and panoptic segmentation.
UPSNet\cite{xiong2019upsnet} proposes a dynamic panoptic head to fuse the results, 
internally mitigates the instance occlusion and prediction conflicts. 
We utilize it as the baseline for its good end-to-end architecture and performance.
Recently, transformer architecture\cite{transformer} has been applied in some panoptic segmentation models\cite{maskformer,max_deeplab}. 
Though achieving preferable performance, these models have a large number of parameters and require huge computing resources for training,
which are difficult to deploy on intelligent vehicles.
Moreover, crop-aware bounding box regression loss and  a novel data sampling and augmentation strategy are proposed in\cite{allscales} for improving panoptic segmentation at all scales.
But they focus more on training strategies and overlook the shortcomings of the model itself.


None of these methods consider the global context information and high-resolution information in the backbone for panoptic segmentation models, 
which are fundamentally required for large-scale things and small-scale stuff. 
For truncated detection of large-scale things, Instance Affinity Head is proposed in \cite{utips} to revise the outputs. 
However, it will bring a considerable amount of parameters and calculations. 
Otherwise, stronger semantic segmentation models are employed in\cite{seamless,chen2020banet,efficientps,allscales}, 
partially improving the performance on small-scale stuff. 
These methods pay attention to the top layers without considering the refined details and the gap between two segmentation branches. 
In this paper, we transfer the attention to underlying layers and make the panoptic segmentation model scale-aware with lightweight modules.

\newpage

\section{Scale-aware Unified Network}
This section will first represent the Pixel-relation Block and the Convectional Network, 
followed by Scale-aware Unified Network (SUNet) for panoptic segmentation.
\subsection{Pixel-relation Block}
\begin{figure}[t]
	\centering
	\includegraphics[width=0.480\linewidth]{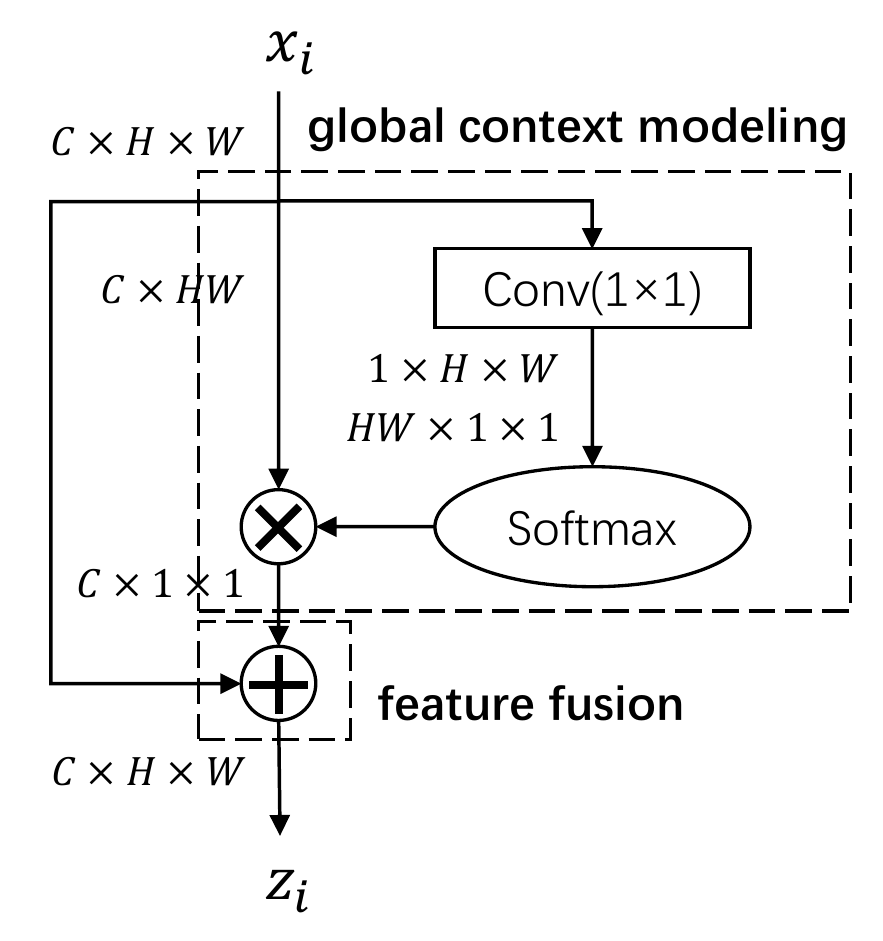}
	\caption{Structure of the Pixel-relation Block. It contains two procedures: (1) global context modeling; (2) feature fusion. It models global context information and long-range dependence for large-scale things.}
	\label{fig:SGCF}
\end{figure}

For the problem of truncated detection on large-scale things, 
global context information and the relationship between pixels are key clues. 
Inspired by Non-local\cite{nonlocal}, Pixel-relation Block is designed for large-scale things detection.

Denote $X={\{x_i\}}_{i=1}^{N_p}$ and $Y={\{y_i\}}_{i=1}^{N_p}$ as input and output feature maps. 
$N_{p}=H \times W$ represents the number of pixels in feature map, 
where $H$ and $W$ are height and width, respectively. 
The general global context modeling operation in neural networks can be defined as:
\begin{equation}
	y_i=\frac{1}{C(x)}\sum_{\forall j}^{}f(x_i,x_j)g(x_j)
	\label{eq:1}
\end{equation}
where $f(\cdot,\cdot)$ is a pairwise function to quantify the ``relationship.'' 
This is a crucial step in global context modeling.
It represents the relationship, such as the affinity between pixel $i$ and $j$,
establishing the long-range dependency between pixels.
Function $g(\cdot)$ is utilized to model the input feature. 
The $g(x_j)$ can be expressed explicitly as $(W_v\cdot x_j)$ in neural networks. 
The $C(x)=\sum_{\forall j}{f(x_i,x_j)}$ is for the normalization. 
Eq. (\ref{eq:1}) enables the output features to be the weighted sum of each position in the input feature map, 
strengthening the relationship between pixels and increasing the receptive field.
Residual learning\cite{resnet} is introduced to avoid model degradation:

\begin{equation}
	z_i = W_zy_i+W_rx_i=W_z\sum_{\forall j}{\frac{f(x_i,x_j)}{C(x)}}(W_v\cdot x_j) +W_r\cdot x_i
	\label{eq:2}
\end{equation}
where $Z={\{z_i\}}_{i=1}^{N_p}$ denotes the output of global context modeling block. 
$W_z$ presents linear transformation and $W_r$ is for shortcut connection, respectively.

Directly introducing this module will bring a substantial computational overhead.
Three steps are adopted in this paper to simplify Eq. (\ref{eq:2}). 
Firstly, considering the discovery in \cite{gcnet}, 
a global attention map can be computed and shared for all query positions. 
More specifically, $\sum_{\forall j}{\frac{f(x_i,x_j)}{C(x)}}(W_v\cdot x_j)$ can be calculated once and utilized for all $x_i$. 
The simplified version of Eq. (\ref{eq:2}) can be expressed as:

\begin{equation}
	z_i = W_z\sum_{\forall j}{\frac{f(x_j)}{C(x)}}(W_v\cdot x_j)+W_r\cdot x_i
	\label{eq:3}
\end{equation}

Secondly, according to the experiment results in \ref{LSB}, 
better performance can be achieved without $W_z$. 
It indicates that $W_z$ plays the role of fully connected layers 
in channel attention\cite{senet} and can be omitted from Eq. (\ref{eq:3}). 
Meanwhile, $W_v$ can also be moved outside the $\sum$ operation using the distributive law. Consequently, Eq. (\ref{eq:3}) can be simplified to: 

\begin{equation}
	z_i = W_v\sum_{\forall j}{\frac{f(x_j)}{C(x)}x_j}+W_r\cdot x_i
	\label{eq:SNL}
\end{equation}

Thirdly, comparing Eq. (\ref{eq:3}) and Eq. (\ref{eq:SNL}), the $W_v$ in Eq. (\ref{eq:SNL}) plays the same role as the $W_z$ in Eq. (\ref{eq:3}). 
So it is intuitive to omit the $W_v$ like $W_z$ in the last step. 
Moreover, take identity mapping$(W_r=I)$ in shortcut connection for simplicity. 
The abstraction of Pixel-relation Block is defined in Eq. (\ref{eq:SGCB}).
$\sum_{\forall j}{\frac{f(x_j)}{C(x)}x_j}$ and $+x_i$ represent global context modeling and feature fusion, respectively.
\begin{equation}
	z_i = \sum_{\forall j}{\frac{f(x_j)}{C(x)}x_j}+x_i
	\label{eq:SGCB}
\end{equation}

Take the prevalent instantiations, Embedded Gaussian\cite{nonlocal}, to model the global but query independent attention. The Pixel-relation Block is shown in Fig.~\ref{fig:SGCF} and defined as:

\begin{equation}
	z_i = \frac{\sum_{j=1}^{N_p}{\exp\left(Wx_j\right)}}{\sum_{n=1}^{N_p}{\exp\left(Wx_n\right)}}x_j + x_i
	\label{eq:SGCB_specified}
\end{equation}

Based on the idea of capturing long-range dependence and modeling global context information, 
Pixel-relation Block is elaborately constructed for large-scale things. 
It can enlarge the receptive fields of pixels and strengthen the connection between pixels, 
which helps panoptic segmentation models to detect large-scale things. 
Moreover, the Pixel-relation Block is obtained through three simplified steps, 
making it lightweight and convenient to be employed in other models. 

\begin{figure}[!t]
	\captionsetup[subfloat]{labelformat=empty, font=scriptsize,labelfont=scriptsize}
	\centering
	\subfloat[(a)]{\includegraphics[width=0.635\linewidth]{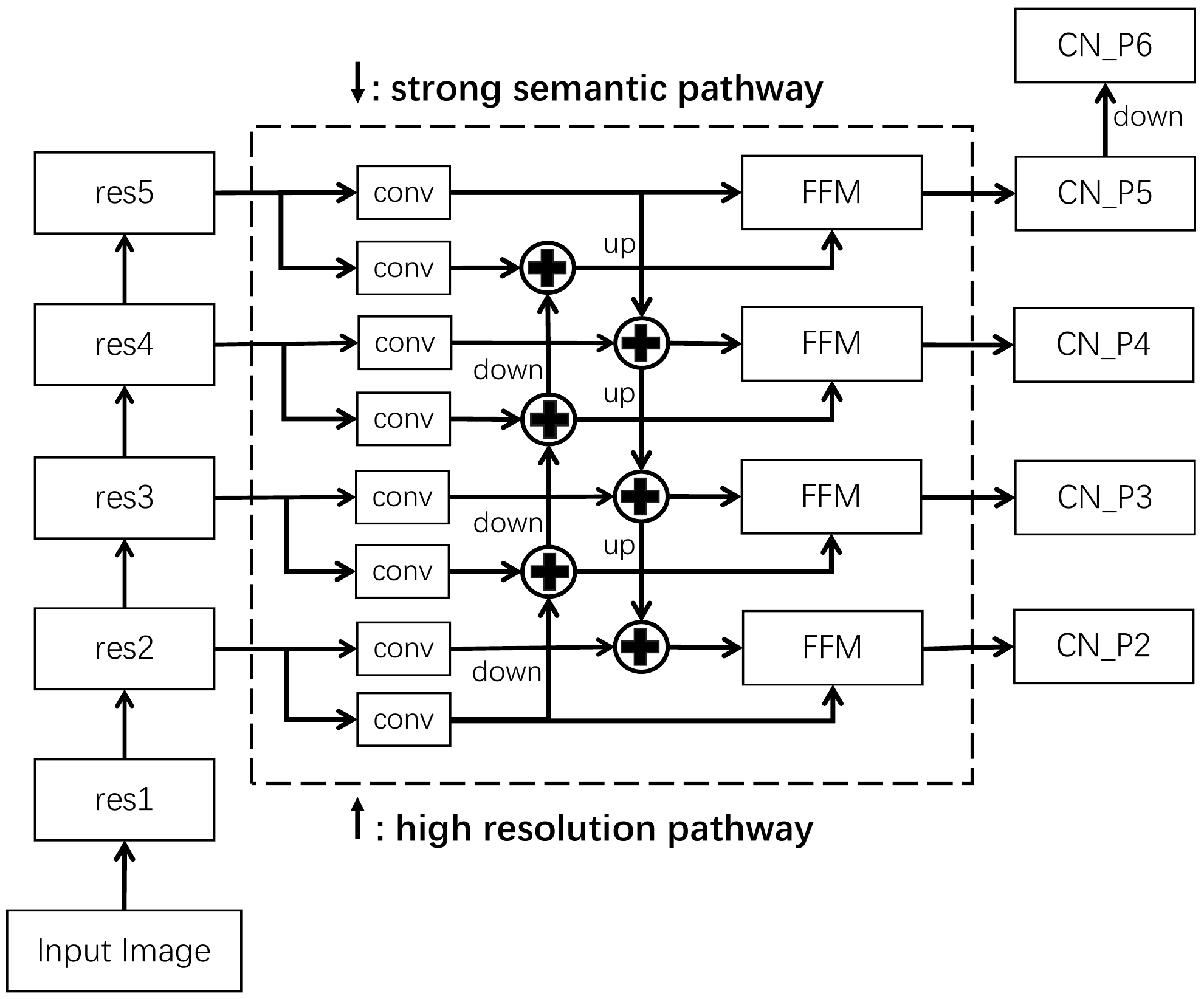}}
	\hspace{0.01cm}
	\subfloat[(b)]{\includegraphics[width=0.285\linewidth]{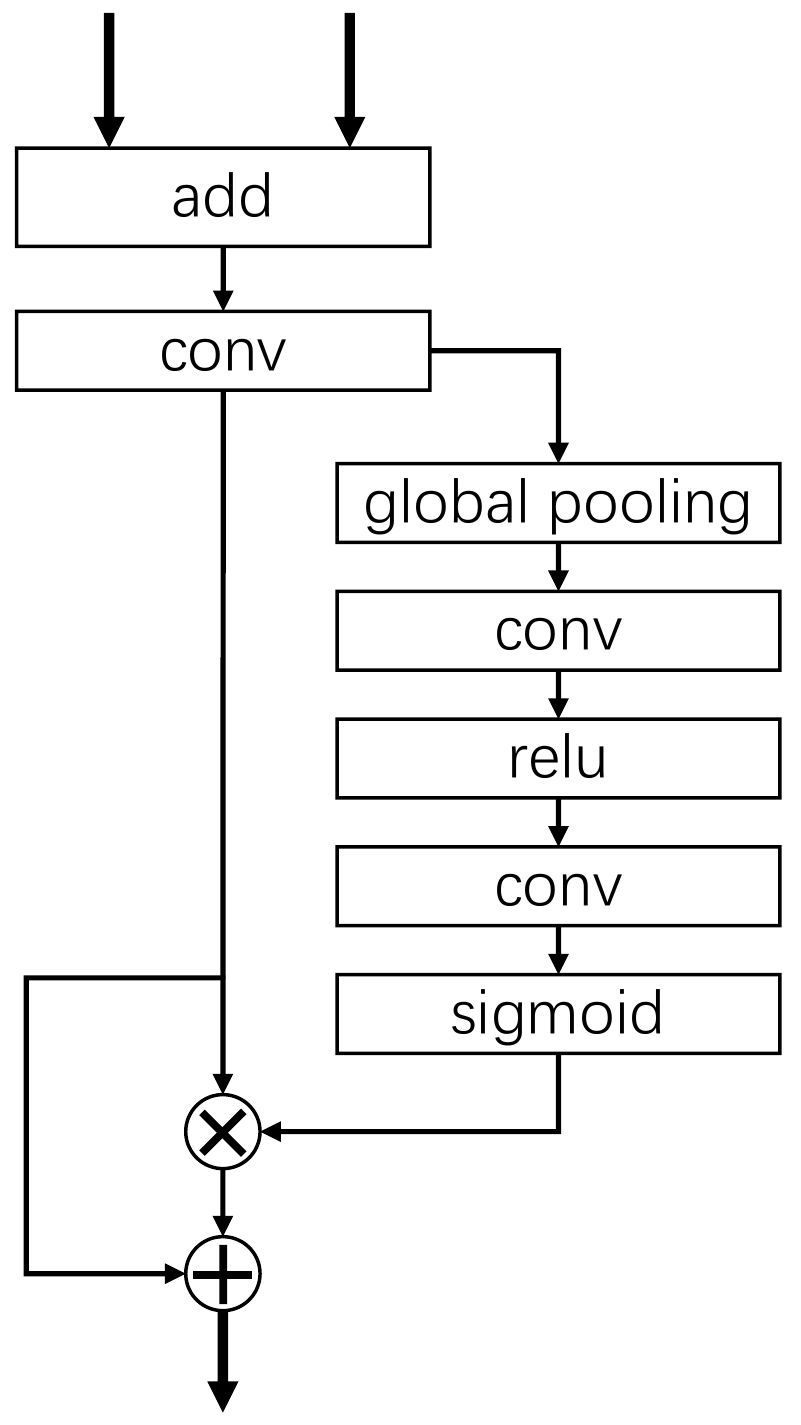}}
	\caption{(a) The structure of Convectional Network. The strong semantic pathway, high resolution pathway, and feature fusion module(FFM) are the three main components. (b) The detailed structure of FFM.}
	\label{fig:Convectional Network}
\end{figure}

\subsection{Convectional Network}

FPN\cite{fpn} has three parts: bottom-up pathway, top-down pathway, and lateral connection. 
The bottom-up pathway extracts useful feature information from images. 
The top-down pathway brings strong semantic features to each layer. 
Features from two pathways are combined via the lateral connection 
and are shared by semantic and instance segmentation branches in panoptic segmentation models. 

FPN is usually used in tasks like object detection and instance segmentation. 
However, there is an additional semantic segmentation branch in the panoptic segmentation model. 
In deep convolutional neural networks, the top-layer features encode the object-level or class-level information with weak spatial clues. 
Contrastively, the low-layer features preserve detailed spatial information like edges and corners. 
For FPN, the information from the top-down pathway degrades the model's performance on small-scale stuff. 
It mainly delivers top-layer features preferred by the instance branch and lacks low-layer information for the semantic branch. 
Additionally, the topdown pathway hallucinates high-resolution features and brings the aliasing effect\cite{fpn}. 
Therefore, it is required to design a new pyramid-like network with more appropriate semantic features for following segmentation task branches.

The detailed structure of the Convectional Network is shown in Fig.~\ref{fig:Convectional Network}(a).  
It consists of three components: strong semantic pathway, high resolution pathway, and feature fusion module (FFM). 
The ``top-to-down'' arrow shows how the strong semantic features flow. 
Meanwhile, the ``down-to-top'' arrow intends to provide high-resolution information. 
It is fundamentally required for segmentation on small-scale stuff. 
FFM is developed to effectively integrate features from two convectional pathways, which is detailed in Fig. \ref{fig:Convectional Network}(b). 
The features from two convectional pathways are first added and go through a $3\times 3$ convolution layer to alleviate the aliasing effect. 
Then channel attention\cite{senet} is utilized to integrate the features. 
Additionally, a shortcut connection is built to ease optimization degradation.
The selection of fusion strategy will be explored and detailed in Section~\ref{CN}.

As a result, the outputs of the Convectional Network have appropriate semantic features and refined details, 
which are more suitable for being shared by the downstream segmentation branches. The outputs are denoted as CN\_P2 to CN\_P6.

\begin{figure*}[!t]
	\centering
	\includegraphics[width=0.8\textwidth]{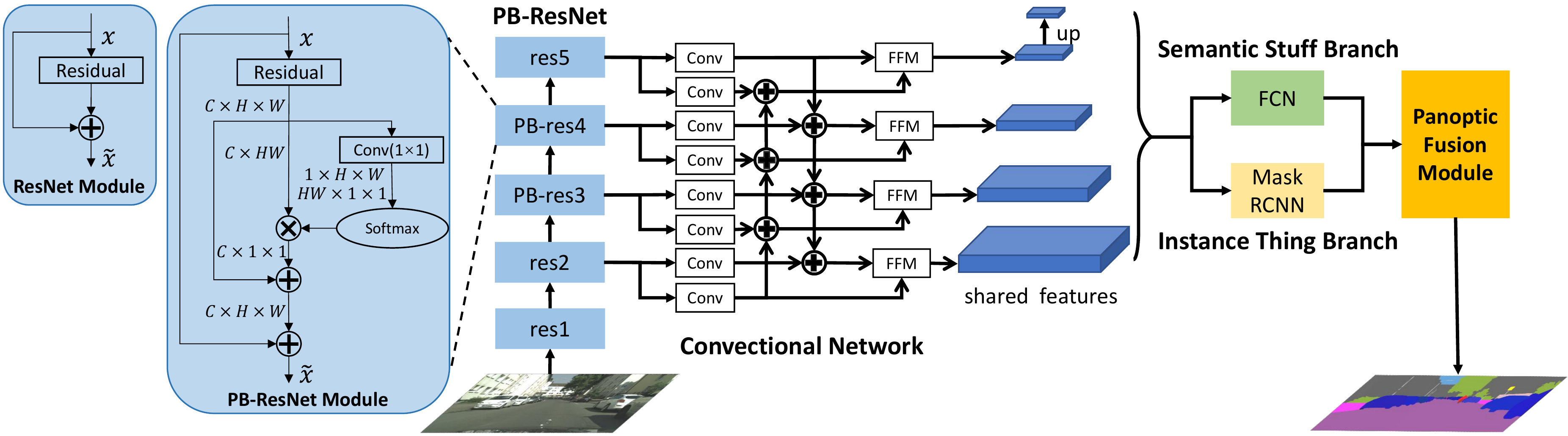}
	\caption{The detailed network structure of Scale-aware Unified Network. 
	PB-ResNet, Convectional Network, Semantic Stuff Branch, Instance Thing Branch, and Panoptic Fusion Module are five main components of SUNet. 
	The Pixel-relation Block and Convectional Network are our main contributions, 
	which make the panoptic segmentation model adapt to multi-scale objects.}
	\label{fig:SUNet}
\end{figure*}

\subsection{Scale-aware Unified Network}
SUNet contains five components: PB-ResNet, Convectional Network, Semantic Stuff Branch, Instance Thing Branch, and Panoptic Fusion Module. The network structure is presented in Fig.~\ref{fig:SUNet}.

\subsubsection{PB-ResNet}
Layers of ResNet are named res1 to res5 for convenience. 
For Pixel-relation Block, 
increased budgets grow with the channels of feature maps, 
while feature maps of deep layers have large receptive fields. 
To make a good trade-off and according to experiments in ~\ref{LSB}, 
Pixel-relation Block is appended to res3 and res4. 
The detailed structure of PB-ResNet Module is presented on the left side of Fig. \ref{fig:SUNet}.

\subsubsection{Convectional Network}
The FPN is replaced by the Convectional Network to provide multiscale features,
which is more appropriate for sharing by following networks.

\subsubsection{Semantic Stuff Branch}

CN\_P2 to CN\_P5 firstly go through two $3\times3$ convolution layers 
and then upsample to the same size of CN\_P2. 
Through concatenation and one convolution layer, 
the output semantic logits are obtained. 
Cross-entropy loss is utilized in the Semantic Stuff Branch.


\subsubsection{Instance Thing Branch}

Mask R-CNN\cite{maskrcnn} is adopted as the Instance Thing Branch. 
The mask logits, corresponding category labels, and bounding boxes are generated for the Panoptic Fusion Module. 
The loss function of Instance Thing Branch is like what in Mask R-CNN\cite{maskrcnn}.

\subsubsection{Panoptic Fusion Module}
The panoptic segmentation head proposed in\cite{xiong2019upsnet} is used  for seamless panoptic segmentation. 
The Panoptic Fusion Module takes the outputs of two branches 
and combines them to the panoptic logits with the channel dimension ($N_{instance}+N_{stuff}+1$). 
$N_{instance}$ and $N_{stuff}$ are the number of detected instance objects and stuff categories, respectively. 
An additional ``unlabel'' layer is appended to alleviate the prediction mistakes. 
The cross-entropy loss is also utilized in the Panoptic Fusion Module.

\subsubsection{Loss Function}
There are three task loss functions in SUNet. 
They are $L_{semantic}$ from Semantic Stuff Branch, $L_{instance}$ from Instance Thing Branch, and $L_{panoptic}$ from the Panoptic Fusion Module. 
The total loss of SUNet is defined in Eq. (\ref{eq:Loss}), where $\lambda_s$, $\lambda_i$, and $\lambda_p$ are hyperparameters for loss balancing during the training process.

\begin{equation}
	Loss = \lambda_sL_{semantic} + \lambda_iL_{instance} + \lambda_pL_{panoptic}
	\label{eq:Loss}
\end{equation}

\section{Experiment}
\subsection{Dataset}
Experiments are conducted on Cityscapes\cite{cityscapes} and COCO\cite{coco} to evaluate the performance of Pixel-relation Block, Convectional Network, and SUNet.

Cityscapes has 5000 images of urban driving scenes with high-quality annotations and 20000 images with coarse annotations. 
We only use the images with fine annotations in our experiments, 
which are divided into 2975, 500, and 1525 for training, validation, and testing. 
The things and stuff categories are 8 and 11 in Cityscapes. 

COCO is a large-scale object detection, segmentation, and captioning dataset, 
covering various scenes in daily life.
It has 80 and 53 classes of things and stuff.
Train2017 and val2017 subsets are used in our experiments, which contain 118k and 5k images for training and validation. 

\subsection{Evaluation Methods and Experiment Settings}
Panoptic quality (PQ) is the combination of segmentation quality (SQ) and recognition quality (RQ):
\begin{small}
\begin{equation}
	PQ = \underbrace{\frac{\sum_{\left(p,g\right)\in{TP}}{IoU\left(p,g\right)}}{|TP|}}_{SQ} \times \underbrace{\frac{|TP|}{|TP|+\frac{1}{2}|FP|+\frac{1}{2}|FN|}}_{RQ}
	\label{eq:PQ_calculation}
\end{equation}
\end{small}
where TP, FP, and FN represent true positives, false positives, and false negatives.
They are used as the evaluation metrics for panoptic segmentation. 
Moreover, mIoU (mean Intersection over Union) and AP (Average Precision) are adopted to evaluate the performance of SUNet on semantic and instance segmentation tasks.

The performance of SUNet is mainly compared with UPSNet\cite{xiong2019upsnet}, which it is based on. 
For a fair comparison, UPSNet is retrained and retested under the same environments as SUNet  
and the reimplemented UPSNet will be presented as UPSNet-r. 
The results reported in \cite{xiong2019upsnet} will also be compared with, which are represented without the -r suffix.
The prediction time of SUNet is averaged on 500 images of the Cityscapes validation set 
with an input resolution of 1024 $\times$ 2048 using a single 1080Ti GPU.

For the software settings, all experiments are conducted using PyTorch in a python environment. 
Learning rate and weight decay are set as 0.0025 and 0.0001 for both datasets.
For Cityscapes, training iterations are 96k and 288k when SUNet takes ResNet50 and ResNet101 as the backbone, respectively.
The learning rate is decayed by a factor of 10 at 72k, 96k, and 192k iterations.
Due to the larger amount of data, the training iterations are 720k for COCO when adopting ResNet50,
and the same learning rate decay is applied at 480k and 640k iterations.
For the coefficients of each term in the loss function (\ref{eq:Loss}), 
$\lambda_{s}$, $\lambda_{i}$, and $\lambda_{p}$ are tuned as 1.5, 1.0, and 0.5 for Cityscapes,
which will be detailed in~\ref{ablation}.
For COCO, they are set as 0.2, 1.0, and 0.1 like\cite{xiong2019upsnet}.

For the hardware settings, two GTX 1080Ti GPUs are used for training and one is used for testing. 
The CPU type is i7-7820X, and 16 cores are utilized in the experiments.

\subsection{Evaluation of Pixel-relation Block}
\label{LSB} 
Pixel-relation Block is gradually appended to res3 to res5 of ResNet to explore where to insert it, 
and the experimental results are shown in Table~\ref{tab:PB_palce}. 
When res3 to res5 are all with Pixel-relation Block, PQ increases from 58.1 to 60.1. 
However, the mIoU degrades, possibly for redundant attention on large-scale things. 
Applying Pixel-relation Block to res3 and res4 brings comparable PQ improvement with higher mIoU and fewer parameters.
Therefore, Pixel-relation Block is added to res3 and res4 in our experiments.

\begin{table}[ht]
	\renewcommand{\arraystretch}{1.0}
	\caption{Results when Pixel-relation Block is added to ResNet.
	\\The best value in each column is highlighted in bold}
	\label{tab:PB_palce}
	\centering
	\resizebox{\linewidth}{0.9cm}{	
	\begin{tabular}{ccc|cccccc}
			\hline
			res3 & res4 & res5 & PQ & SQ & RQ & AP & mIoU &Params(M) \\ \hline
			& & & 58.1 & 79.5 & 71.7 & 32.6 & 75.0 &\bfseries{44.085} \\ 
			$\surd$ & & & 59.1 & 80.2 & 72.4 & \bfseries{33.9} & \bfseries{77.2} &44.087 \\ 
			$\surd$ & $\surd$ & & 60.0 & 80.2 & 73.5 & 33.5 & 76.4 & 44.093 \\ 
			$\surd$ & $\surd$ & $\surd$ & \bfseries{60.1} & \bfseries{80.3} & \bfseries{73.6} & 33.5 & 74.8 &44.100 \\ \hline
	\end{tabular}}

\end{table}

\begin{figure}[!ht]
	\centering
		\includegraphics[width=0.49\linewidth]{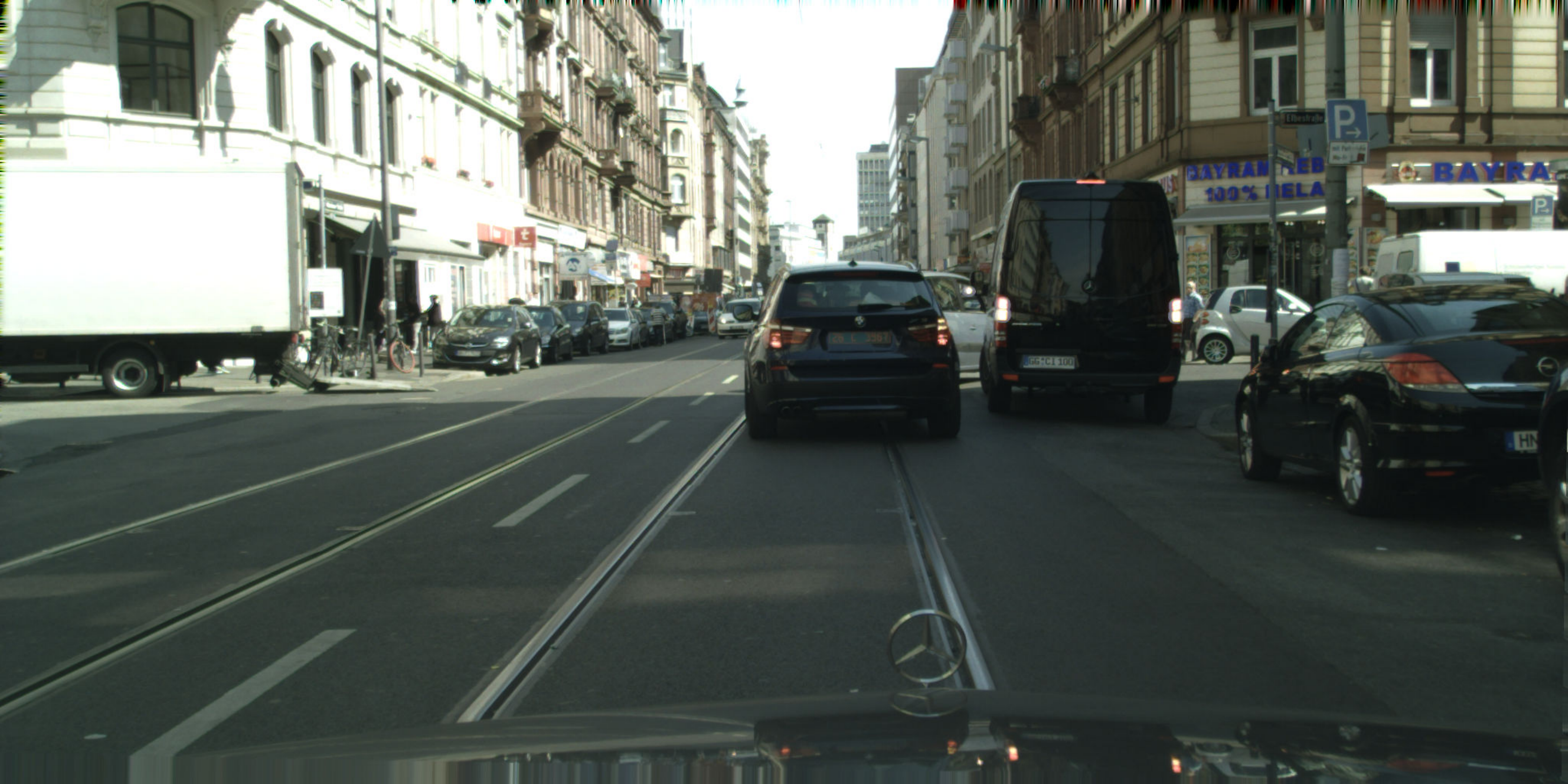}
		\includegraphics[width=0.49\linewidth]{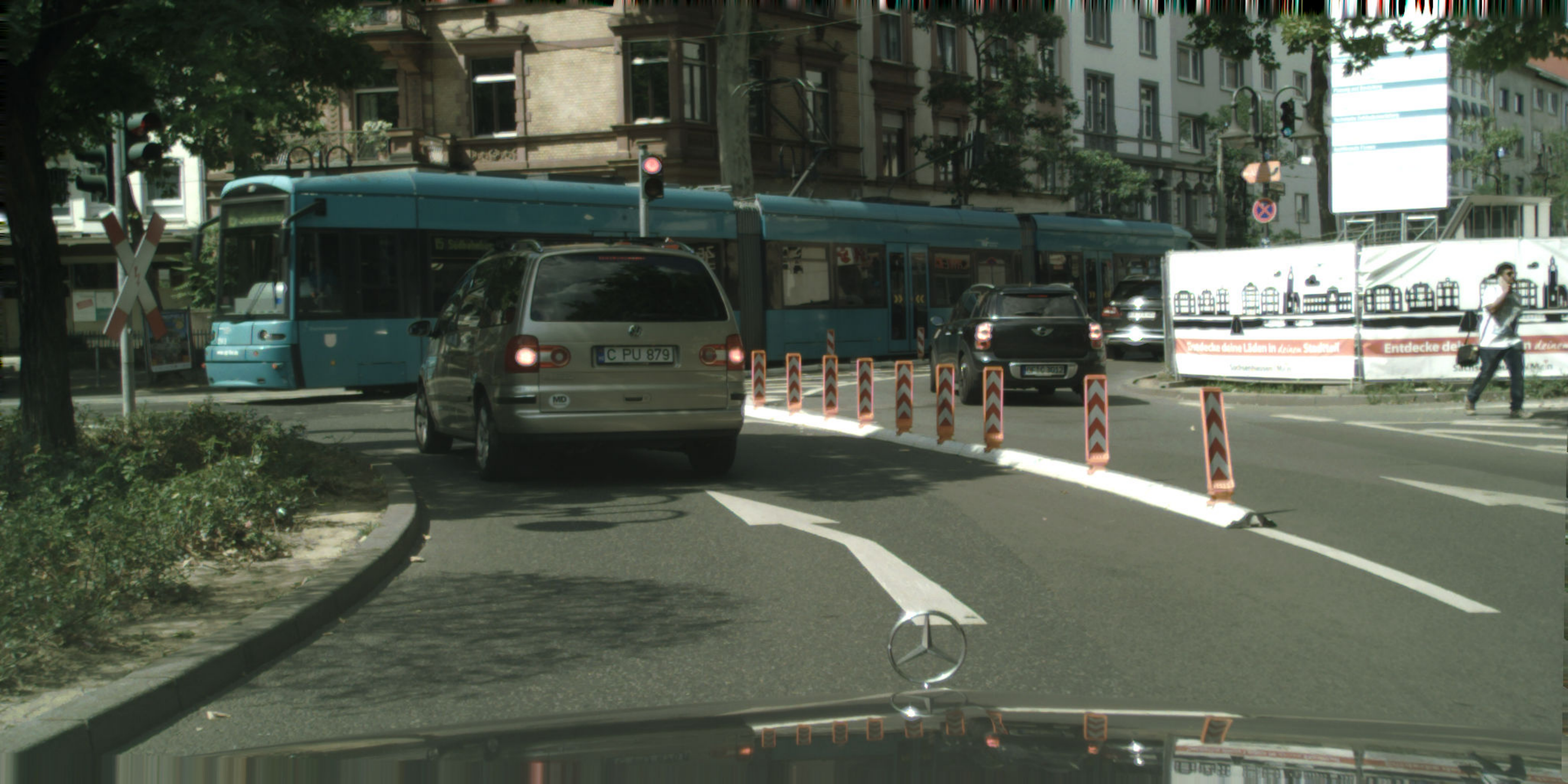} \\ \vspace{0.05cm}
		\includegraphics[width=0.49\linewidth]{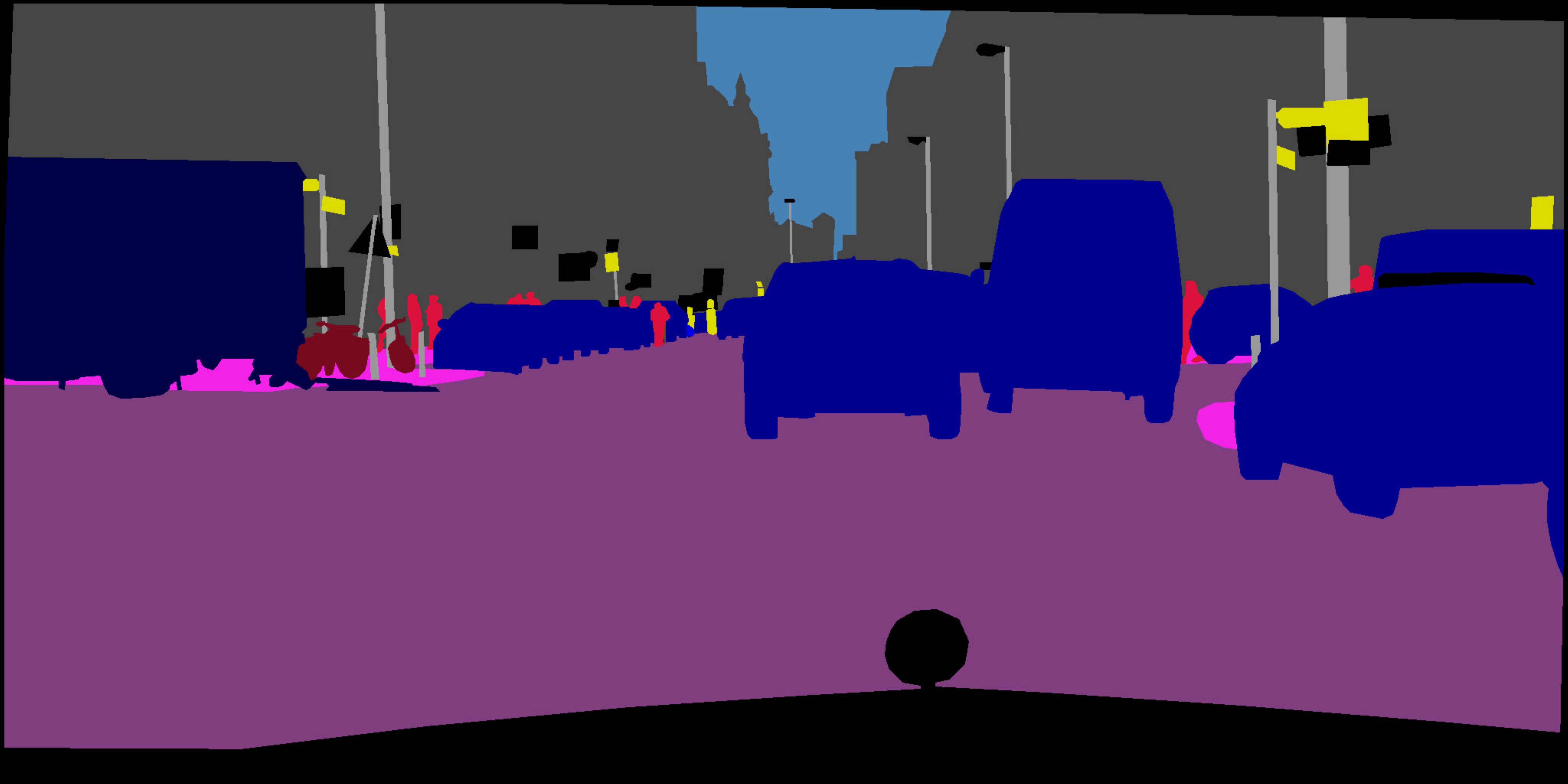}
		\includegraphics[width=0.49\linewidth]{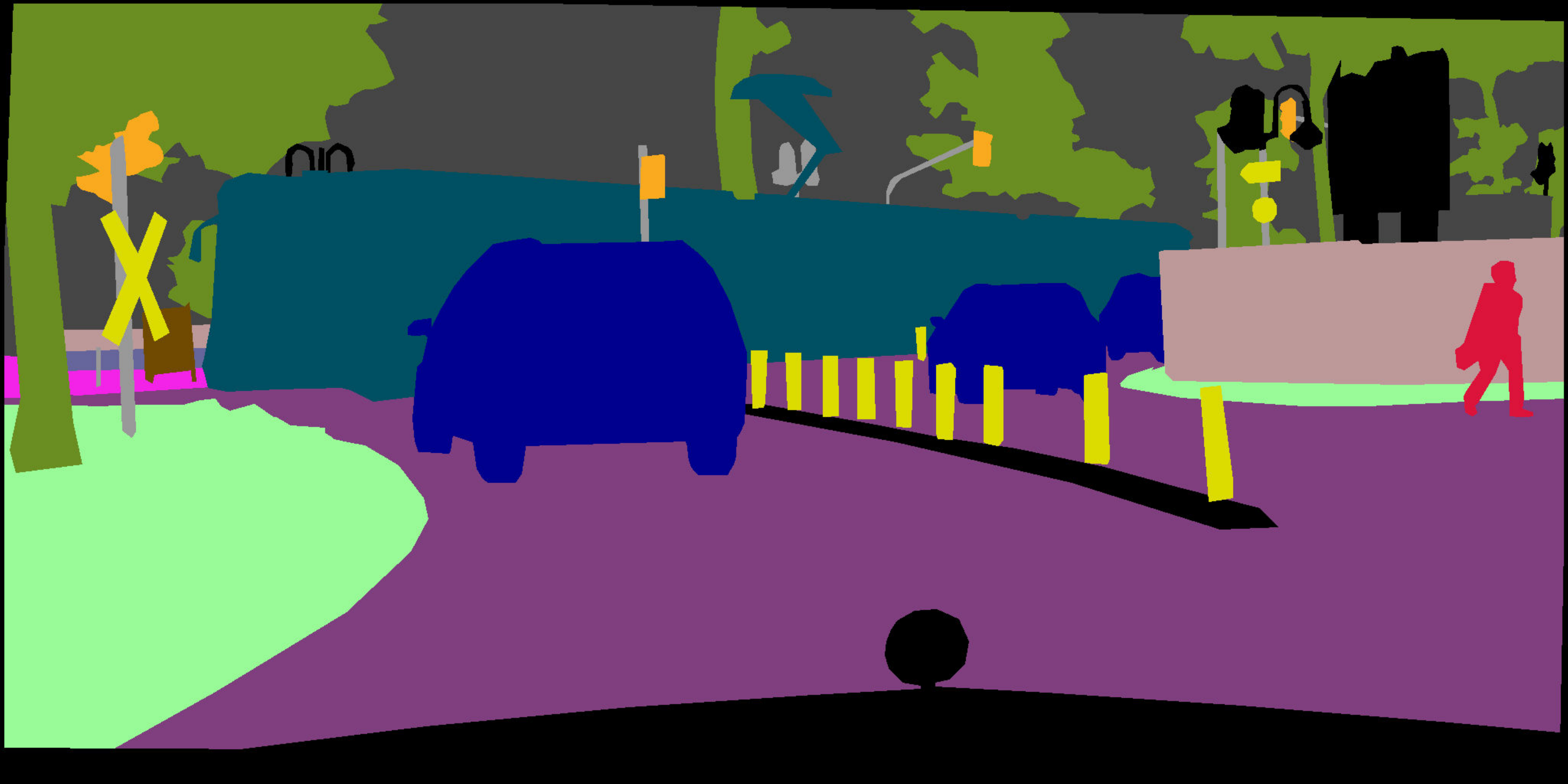} \\ \vspace{0.05cm}
		\includegraphics[width=0.49\linewidth]{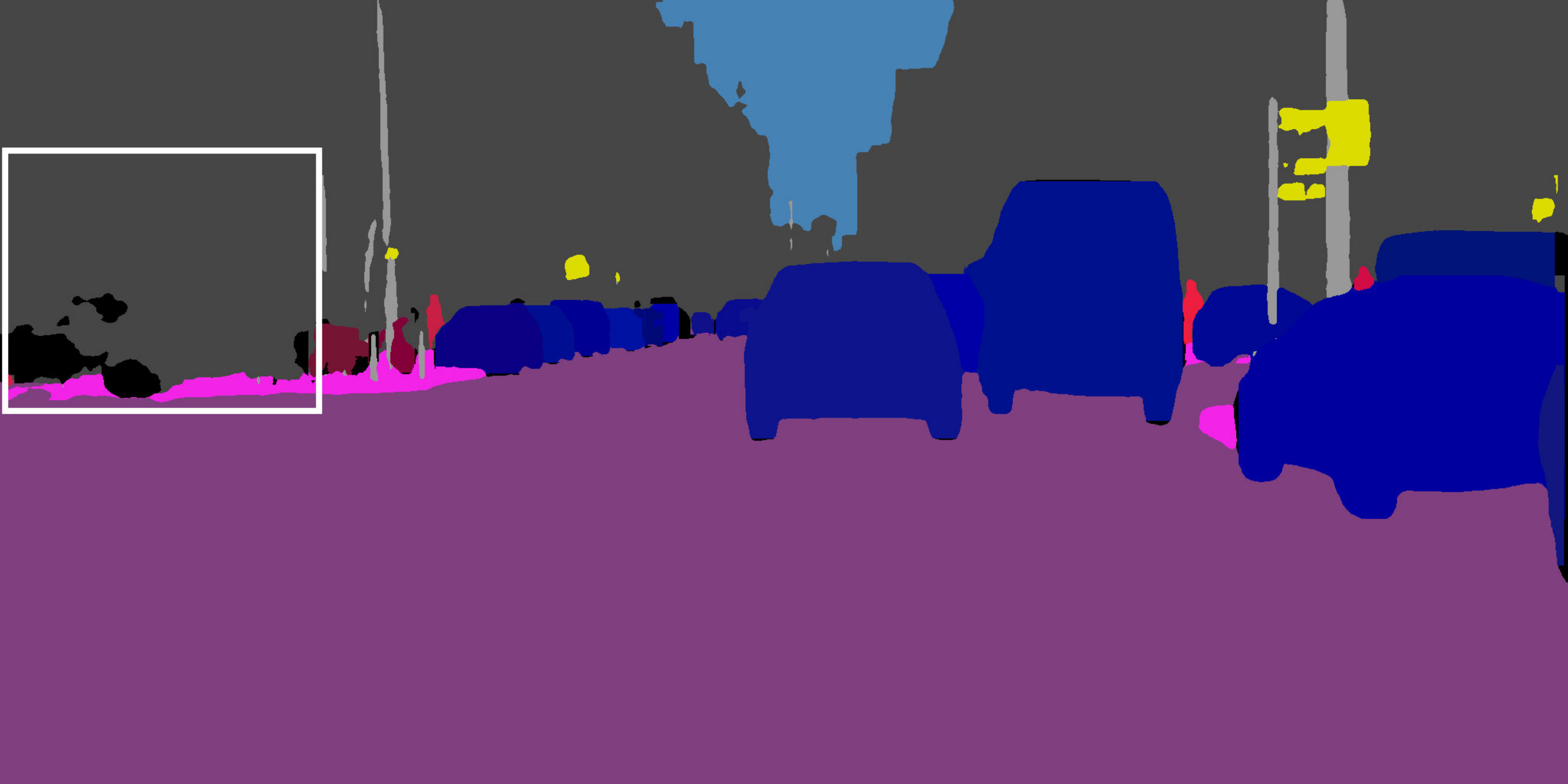}
		\includegraphics[width=0.49\linewidth]{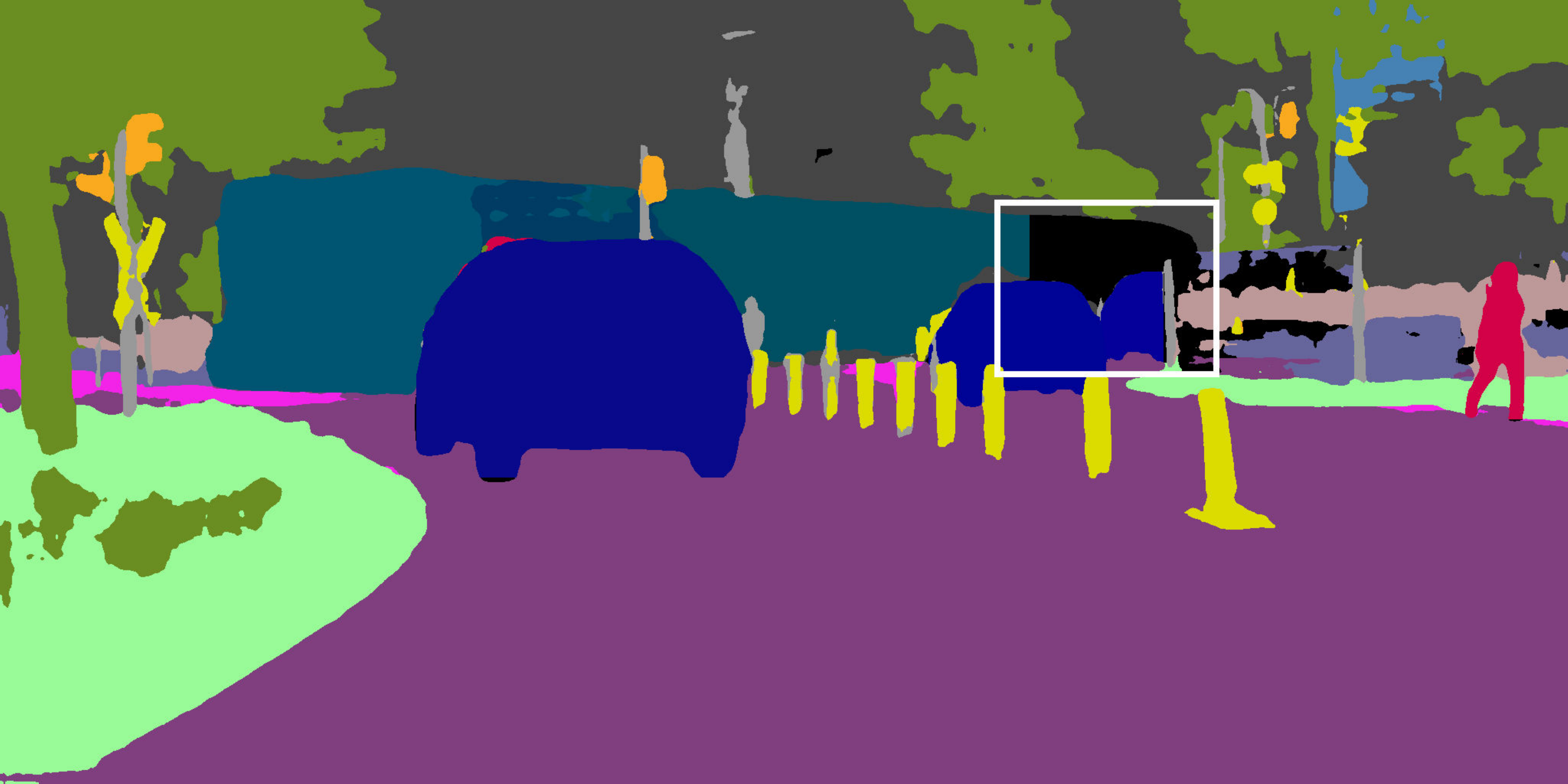} \\ \vspace{0.05cm}
		\includegraphics[width=0.49\linewidth]{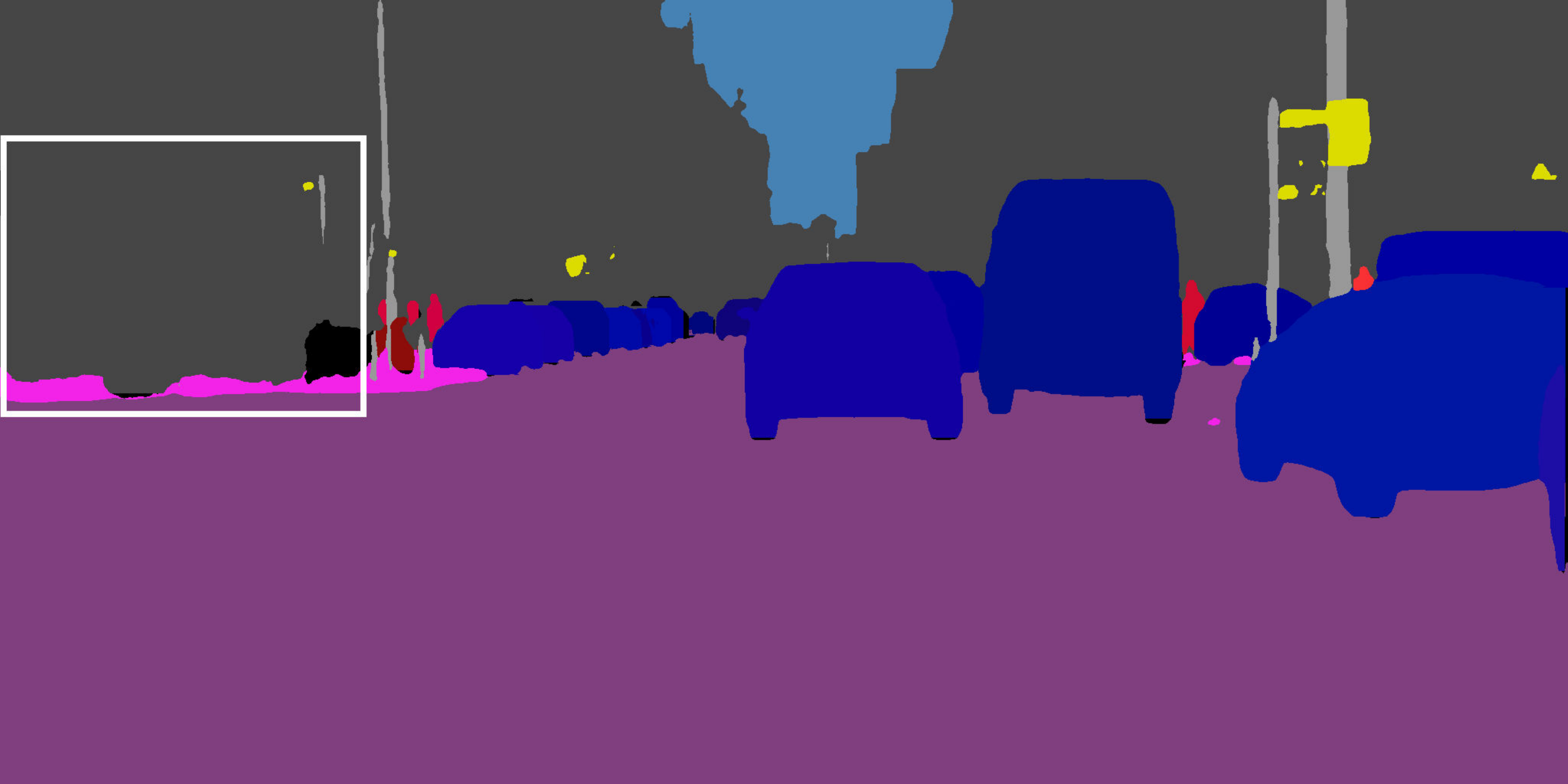}
		\includegraphics[width=0.49\linewidth]{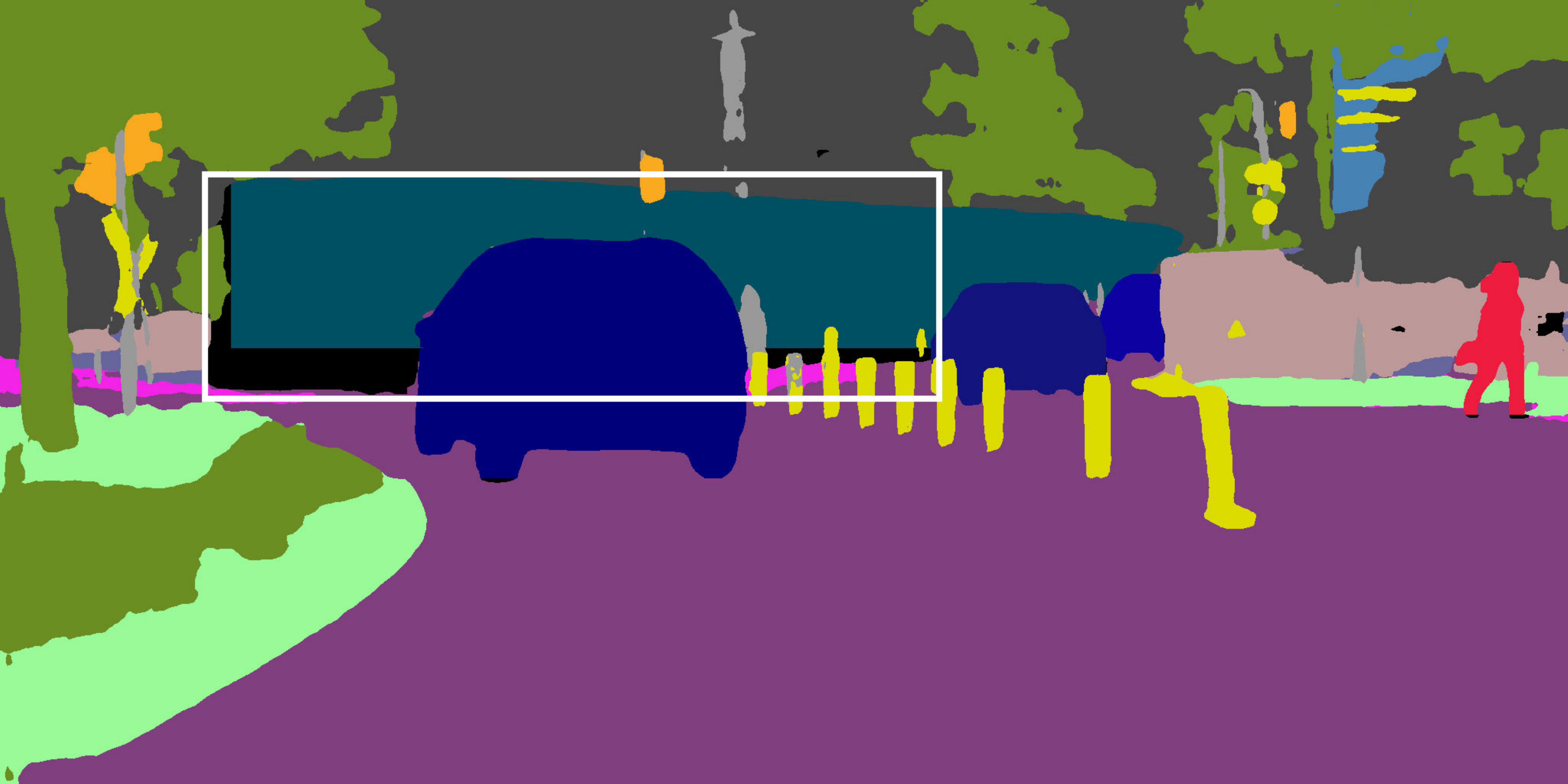} \\ \vspace{0.05cm}
		\includegraphics[width=0.49\linewidth]{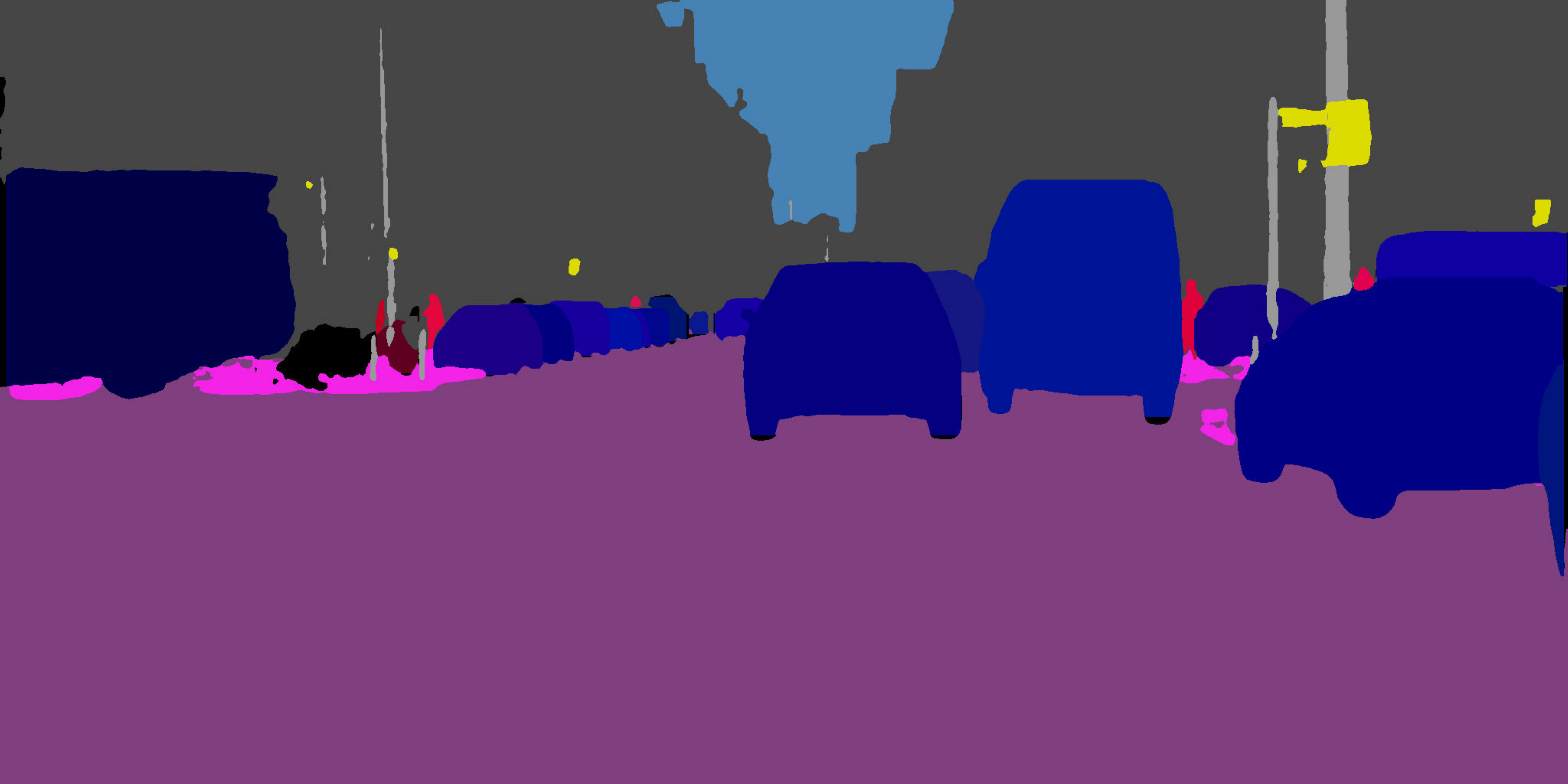}
		\includegraphics[width=0.49\linewidth]{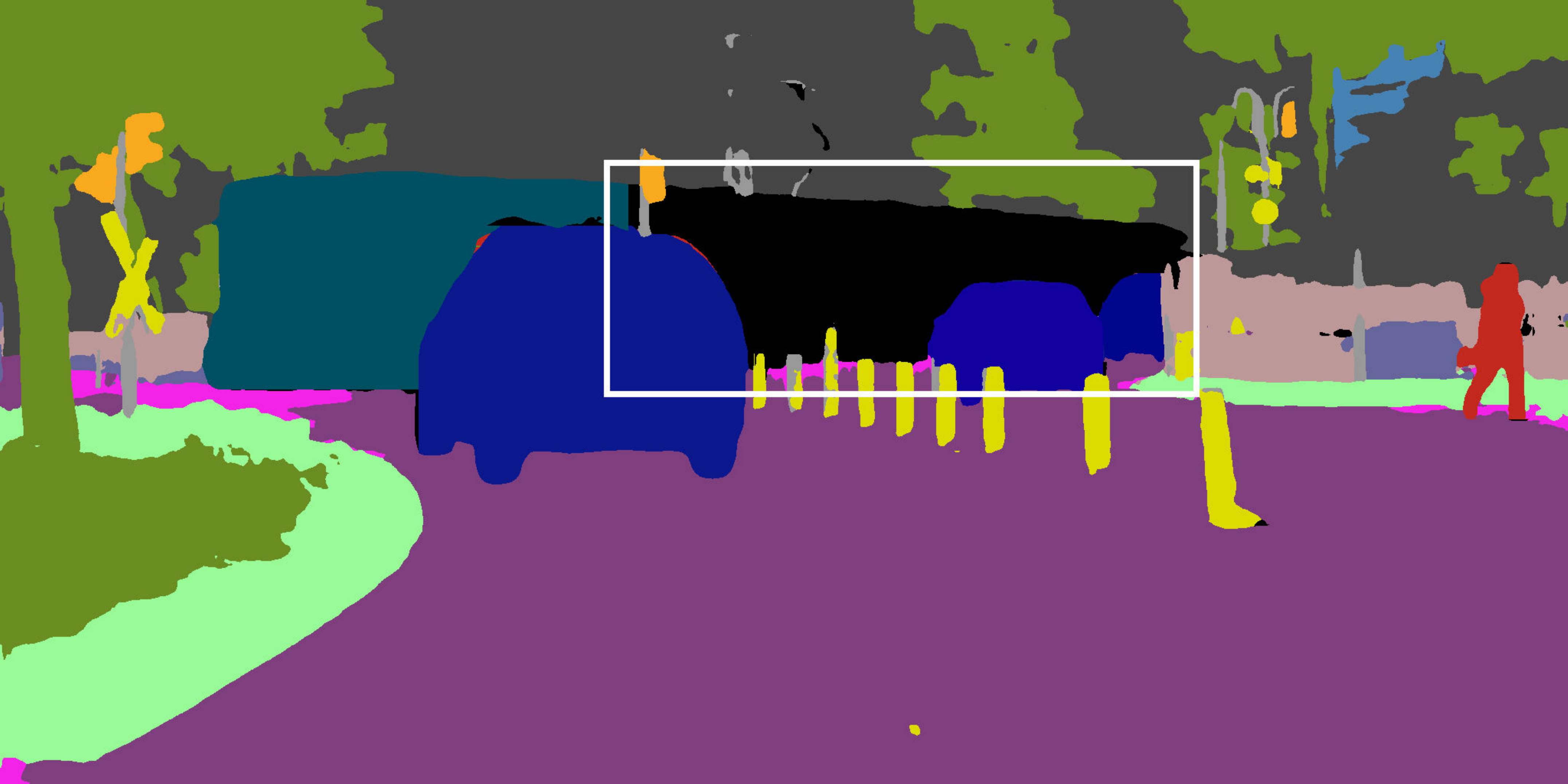} \\ \vspace{0.05cm}
		\includegraphics[width=0.49\linewidth]{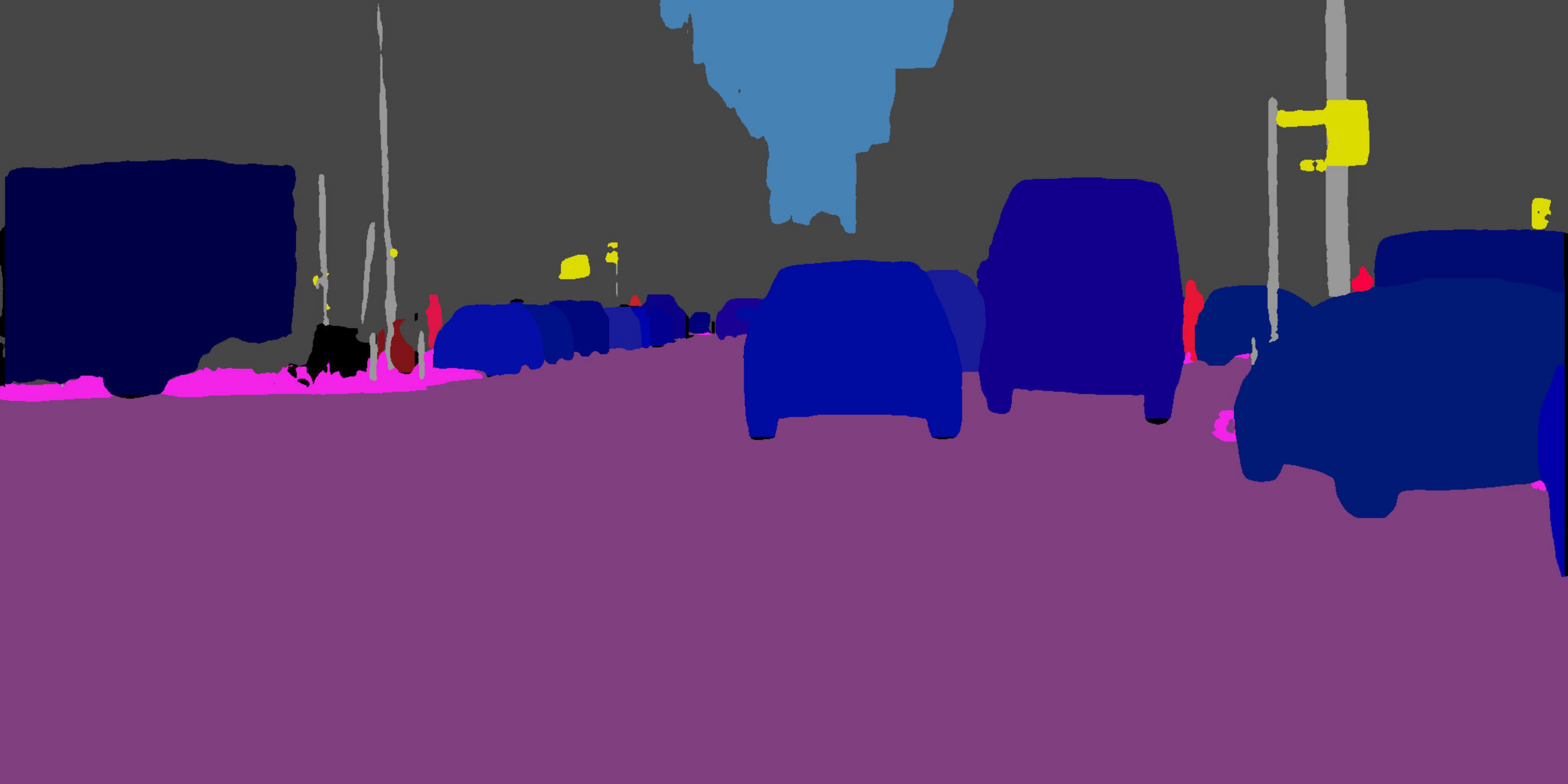}
		\includegraphics[width=0.49\linewidth]{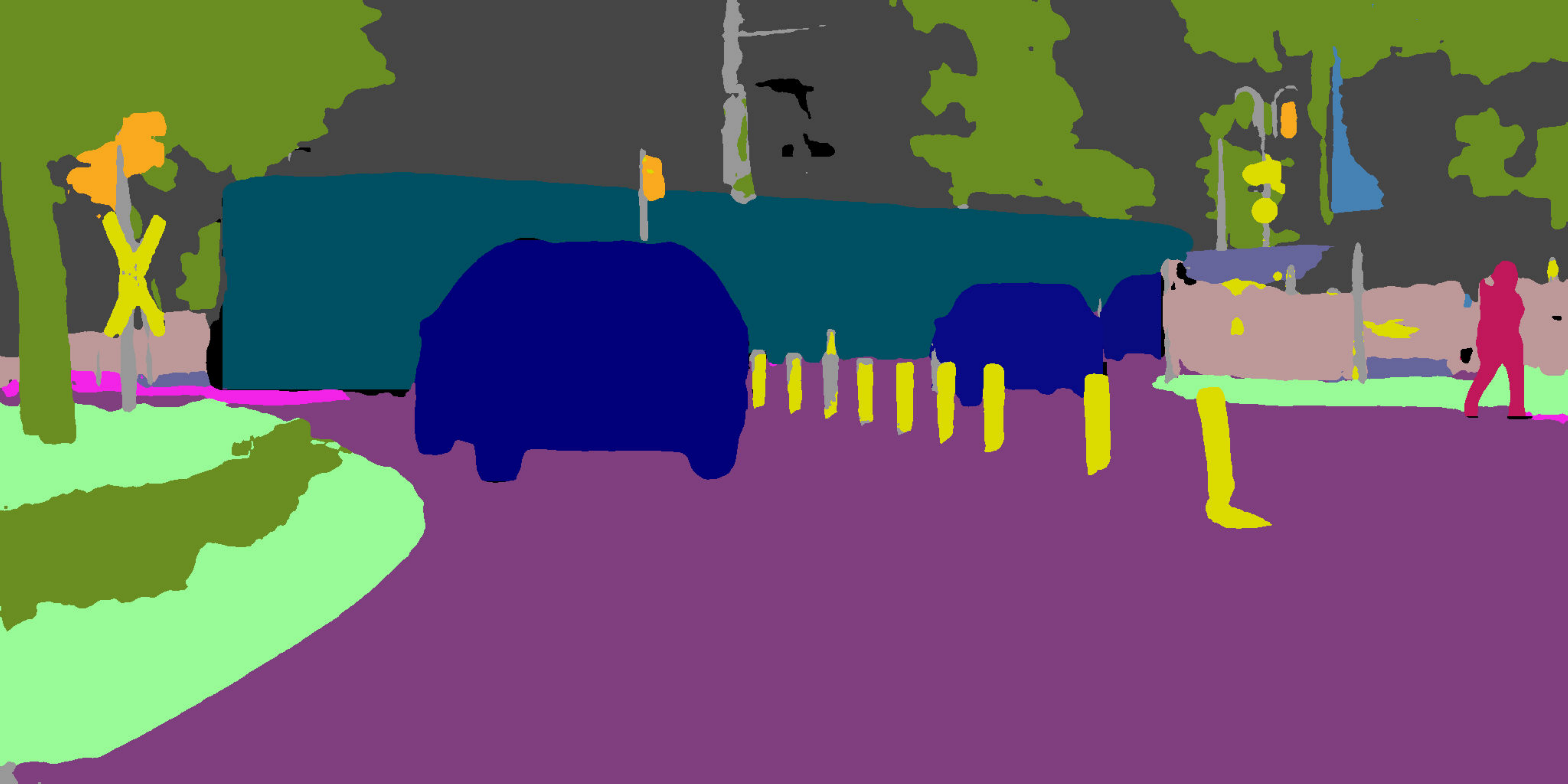} \\ 
	\caption{Qualitative segmentation results of panoptic segmentation models with different context modeling blocks. 
	Problems are marked by white boxes for better viewing.
	From top to down are raw images, ground truth, predictions of UPSNet50-r, UPSNet50-r with SE Block, CBAM Block, and Pixel-relation Block, respectively.
	Pixel-relation Block helps model segment large-scale things.
	}
	\label{fig:SGCB_result}
\end{figure}

Pixel-relation Block is also compared with two classical context modeling modules: 
SE Block\cite{senet} and CBAM Block\cite{cbam}. 
SE Block has a channel attention mechanism, while
CBAM Block\cite{cbam} introduces both channel and spatial attention. 
They are integrated into res3 and res4 for a fair comparison, and 
the comparison results are shown in Table~\ref{tab:SGCB}. 

When res3 and res4 of ResNet50 are with Pixel-relation Block, PQ of all, things, and stuff categories increased by 1.9, 2.3, and 1.6, respectively. 
SQ and RQ are improved by 0.7 and 1.8. 
AP and mIoU are also with  0.9 and 1.4 increments. 
These results demonstrate that the Pixel-relation Block enhances the model's performance, especially on things categories. 
Meanwhile, Pixel-relation Block surpasses the other two modules with a healthy margin and brings fewer parameters,
only increasing the prediction time of 8.1 ms.

The qualitative panoptic segmentation results are shown in Fig.~\ref{fig:SGCB_result}. 
From top to down are raw images, ground truth, 
predictions of UPSNet50-r, UPSNet50-r with SE Block, CBAM Block,
and Pixel-relation Block, respectively.
Pixel-relation Block improves the segmentation results on large-scale things and performs better than the other two modules. 


Moreover, we append Pixel-relation Block to res3 and res4 of ResNet50 in Panoptic FCN\cite{panopticfcn} to explore the generalization of it, 
which is shown in Table~\ref{tab:PB_FCN}.
With Pixel-relation Block, Panoptic FCN gains 2.2, 0.9, 2.1, 1.8, and 0.9 improvements on PQ, SQ, RQ, AP, and mIoU, respectively, 
with only 0.953 M additional parameters.
The steady improvement demonstrates the effectiveness of this module.

\subsection{Evaluation of Convectional Network}
\label{CN}
To explore the strategy to fuse the information from the two flows in Convectional Network, 
addition, concatenation, and channel attention are adopted in experiments. 
The experimental results are presented in Table~\ref{tab:convec abla}. 

\begin{table*}[!ht]
	\renewcommand{\arraystretch}{1.0}
	\caption{Performance and comparison of Pixel-relation Block and Convectional Network on Cityscapes validation set.\\The best value in each column is highlighted in bold}
	\label{tab:SGCB}
	\centering
	\begin{tabular}{c|c|c|c|c|c|c|c|c|c|c|c|c|c}
		\hline
		\multirow{2}{*}{method} & \multicolumn{3}{c|}{PQ} & \multicolumn{3}{c|}{SQ} & \multicolumn{3}{c|}{RQ} & AP&mIoU& \multirow{2}{*}{Params(M)}&\multirow{2}{*}{Time(ms)} \\ \cline{2-12} 
		&all	&things	&stuff	&all	&things	&stuff	&all	&things 	&stuff 	& all & all &	&  \\ \hline
		UPSNet-r (baseline)	&58.1	&52.0	&62.5	&79.5	&78.7	&80.1	&71.7	&65.9	&75.9	& 32.6 & 75.0 &\bfseries{44.085} &\bfseries{275.6}\\ \hline \hline
		+SE Block\cite{senet}	&57.8	& 52.8 & 61.5 & 79.5 & 79.4 & 79.6 & 71.3 & 66.3 & 74.9 &  32.0 & 75.3 & 45.003 &303.5\\ \hline
		+CBAM Block\cite{cbam}  &58.3 &52.9 &62.2 &79.5 & 79.1 &79.7 &71.9 &66.5 &75.8 &32.6 &75.7 & 45.004 & 381.5\\ \hline
		+\bfseries{Pixel-relation Block}	&\bfseries{60.0}	&\bfseries{54.3}	&\bfseries{64.1}	&\bfseries{80.2}	&\bfseries{79.2}	&\bfseries{80.9}	&\bfseries{73.5}	&\bfseries{68.2}	&\bfseries{77.3}	& \bfseries{33.5} & \bfseries{76.4} &44.093 &283.7\\
		\hline \hline
		+2-way FPN\cite{efficientps} & 59.1 & 53.0 & 63.6 & 80.2 & 79.7 & 80.6 & 72.4 & 66.2 & 76.8 & 33.8 & 76.9 & 45.069 &279.5\\ \hline
		+\bfseries{Convectional Network}	&\bfseries{60.1}	&\bfseries{54.9}	&\bfseries{63.9}	&\bfseries{80.4}	&\bfseries{79.7}	&\bfseries{80.9}	&\bfseries{73.4}	&\bfseries{68.6}	&\bfseries{76.9}	&\bfseries{34.3}	&\bfseries{77.2} & 45.077 &295.9\\ \hline
	\end{tabular}
\end{table*}

\begin{table}[ht]
	\renewcommand{\arraystretch}{1.0}
	\caption{Experimental Results of different fusion types for Convectional Network. 
	Add, concat and attention represent for addition, concatenation and channel attention.
	\\The best value in each column is highlighted in bold}
	\label{tab:convec abla}
	\centering
	\resizebox{\linewidth}{0.9cm}{	
		\begin{tabular}{ccc|cccccc}
			\hline
			add & concat & attention & PQ & SQ & RQ & AP & mIoU &Params(M) \\ \hline
			& & & 58.1 & 79.5 & 71.7 & 32.6 & 75.0 &\bfseries{44.085} \\ 
			$\surd$ & & & 59.1 & 80.2 & 72.4 & 33.8 & 76.9 &45.069 \\ 
			$\surd$ &  & $\surd$ & 60.1 & 80.4 & 73.4 & \bfseries{34.3} & \bfseries{77.2} & 45.077 \\ 
			 & $\surd$ & $\surd$ & \bfseries{60.2} & \bfseries{80.5} & \bfseries{73.5} & 34.2 & 75.6 &47.437 \\ 
			 \hline
	\end{tabular}}
\end{table}

The simple addition of convectional features brings 1.0 improvement in PQ, 
and subsequent attention fusion brings an additional 1.0 improvement.
Compared with addition, concatenation only brings a 0.1 increase in PQ but has more 2.36 M parameters.
Therefore, the combination of addition and channel attention is adopted for effectiveness and lightweight.

The performance of Convectional Network on Cityscapes validation set is shown in Table~\ref{tab:SGCB}. 
With Convectional Network, there are 2.0, 0.9, 1.7, 1.7, and 2.2 improvements in PQ, SQ, RQ, AP, and mIoU, respectively.
It implies that features supplied by Convectional Network are more suitable for being shared by the downstream segmentation branches. 
Compared with 2-way FPN proposed in \cite{efficientps}, 
Convectional Network performs better for its effective feature fusion mechanism,
accompanied by a 2.6 ms increase in prediction time.

\begin{figure}[!ht]
	\centering
		\includegraphics[width=0.49\linewidth]{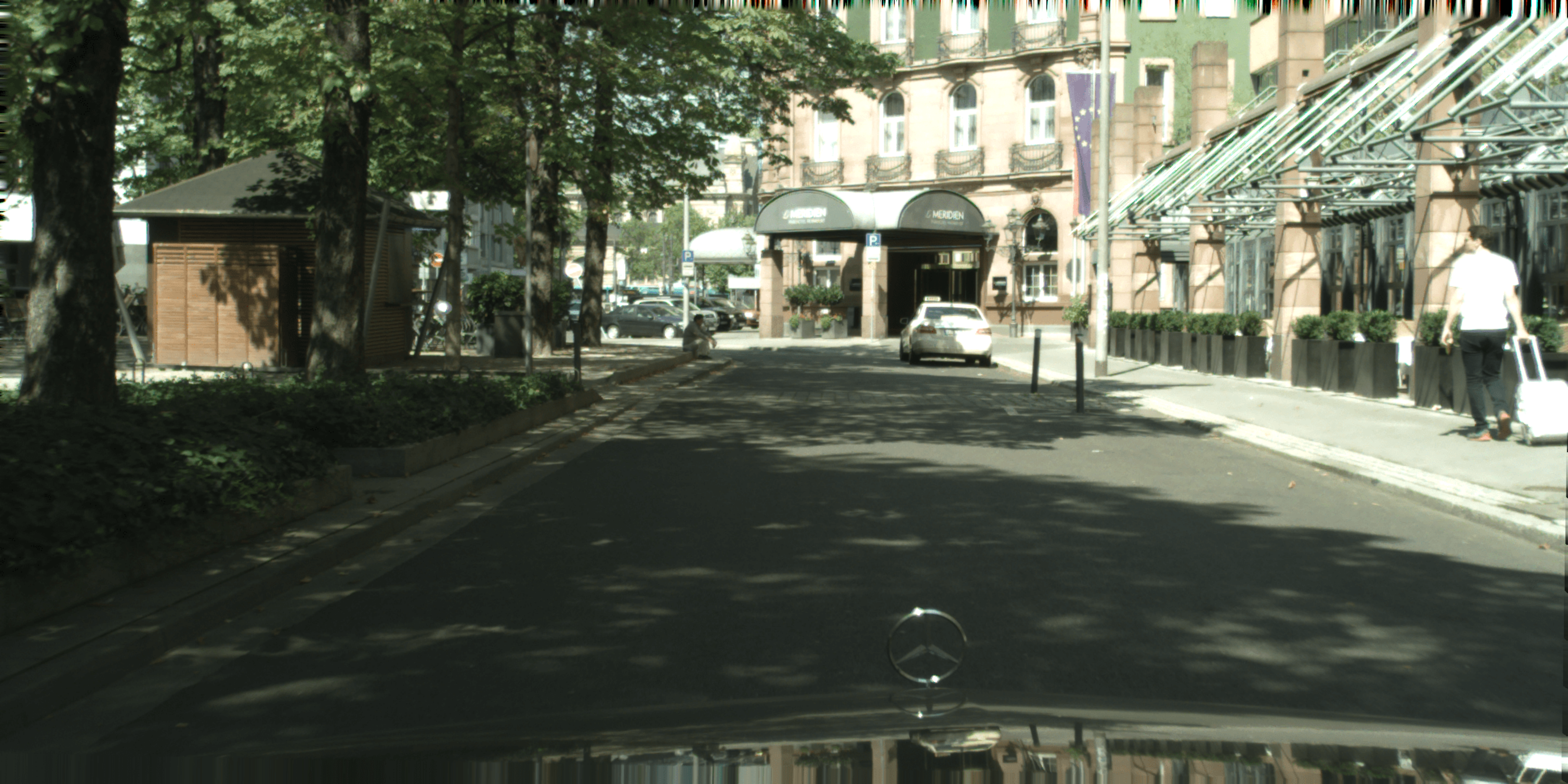}
		\includegraphics[width=0.49\linewidth]{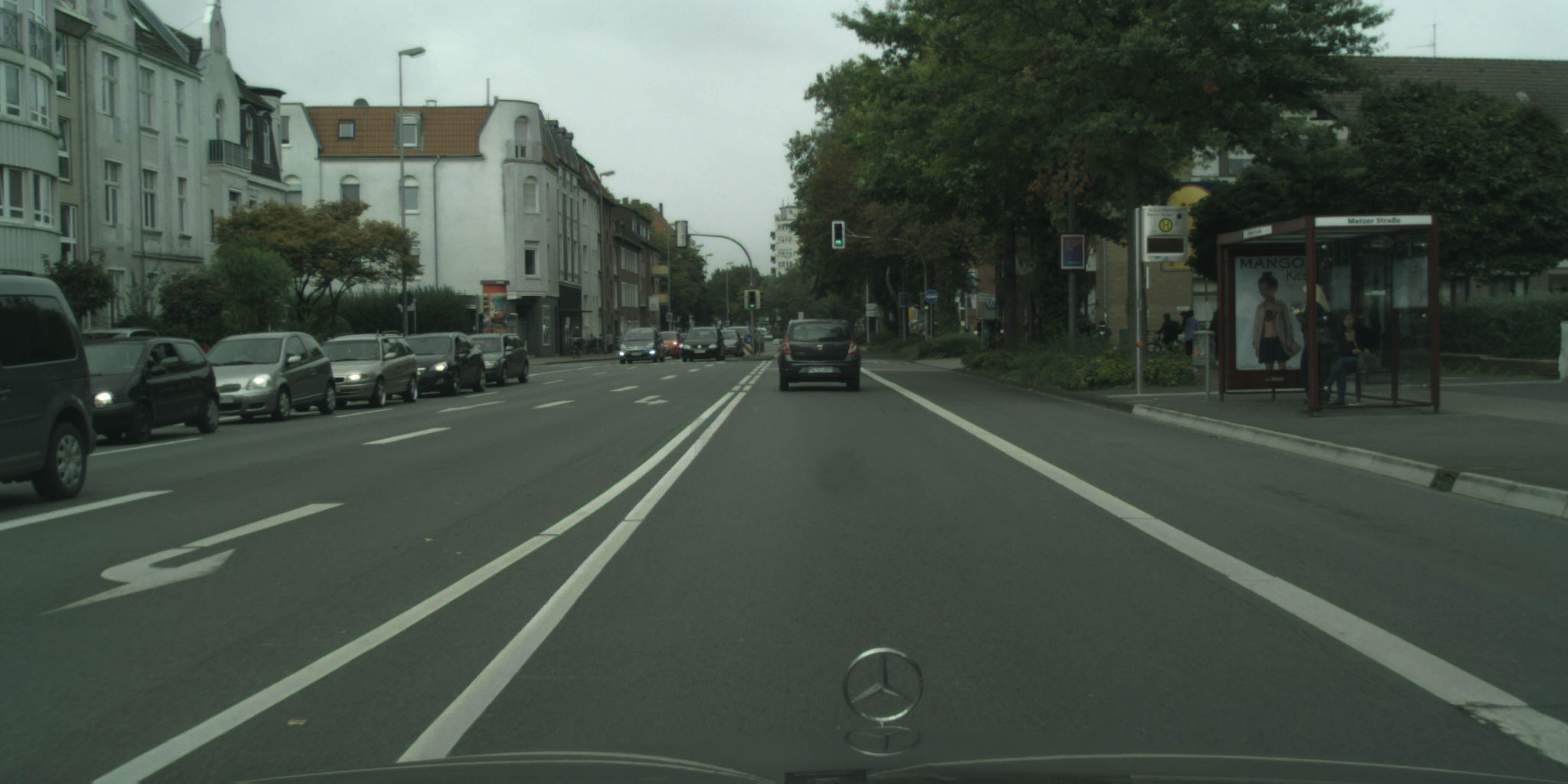}\\ \vspace{0.05cm}
		\includegraphics[width=0.49\linewidth]{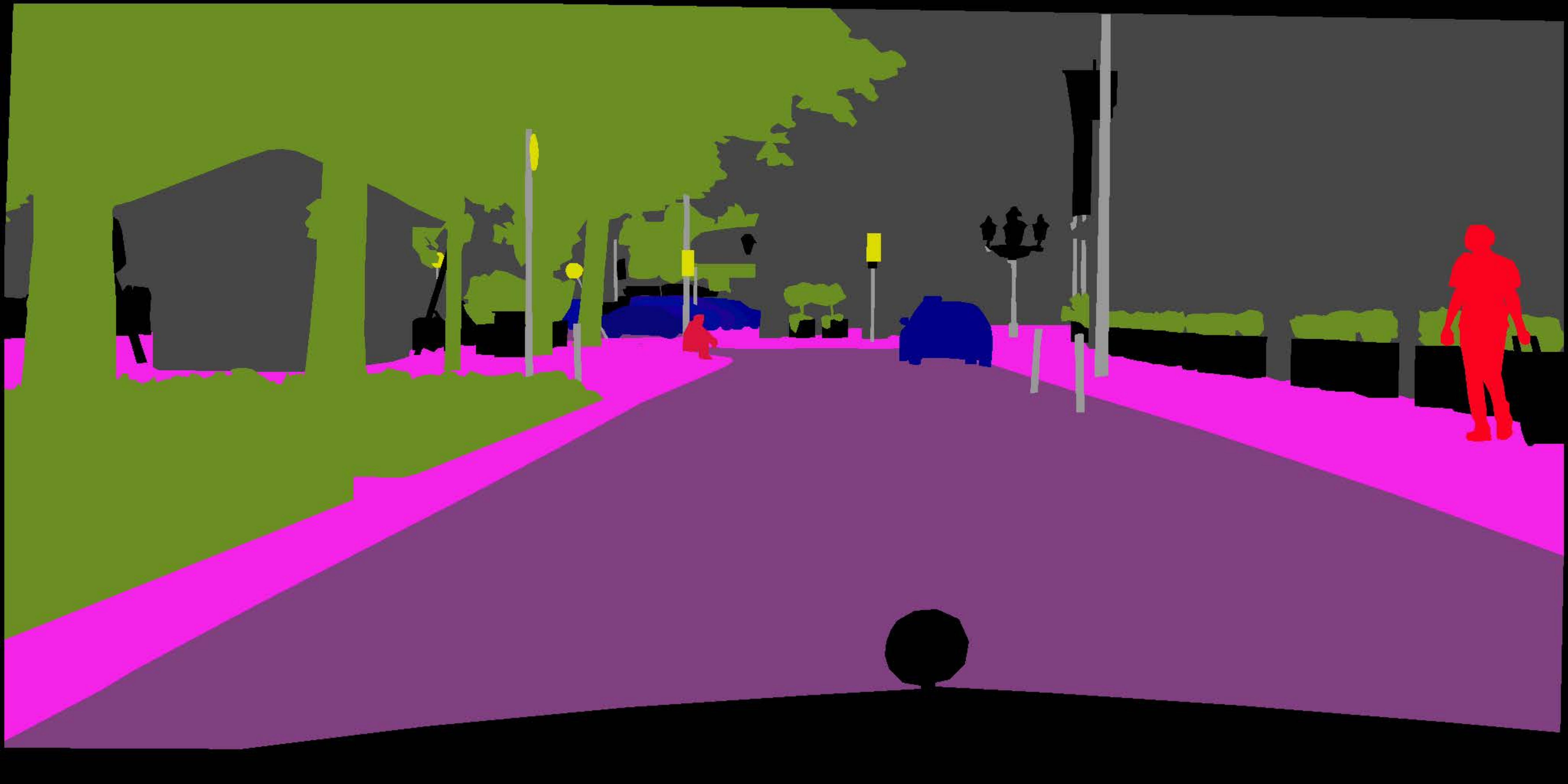}
		\includegraphics[width=0.49\linewidth]{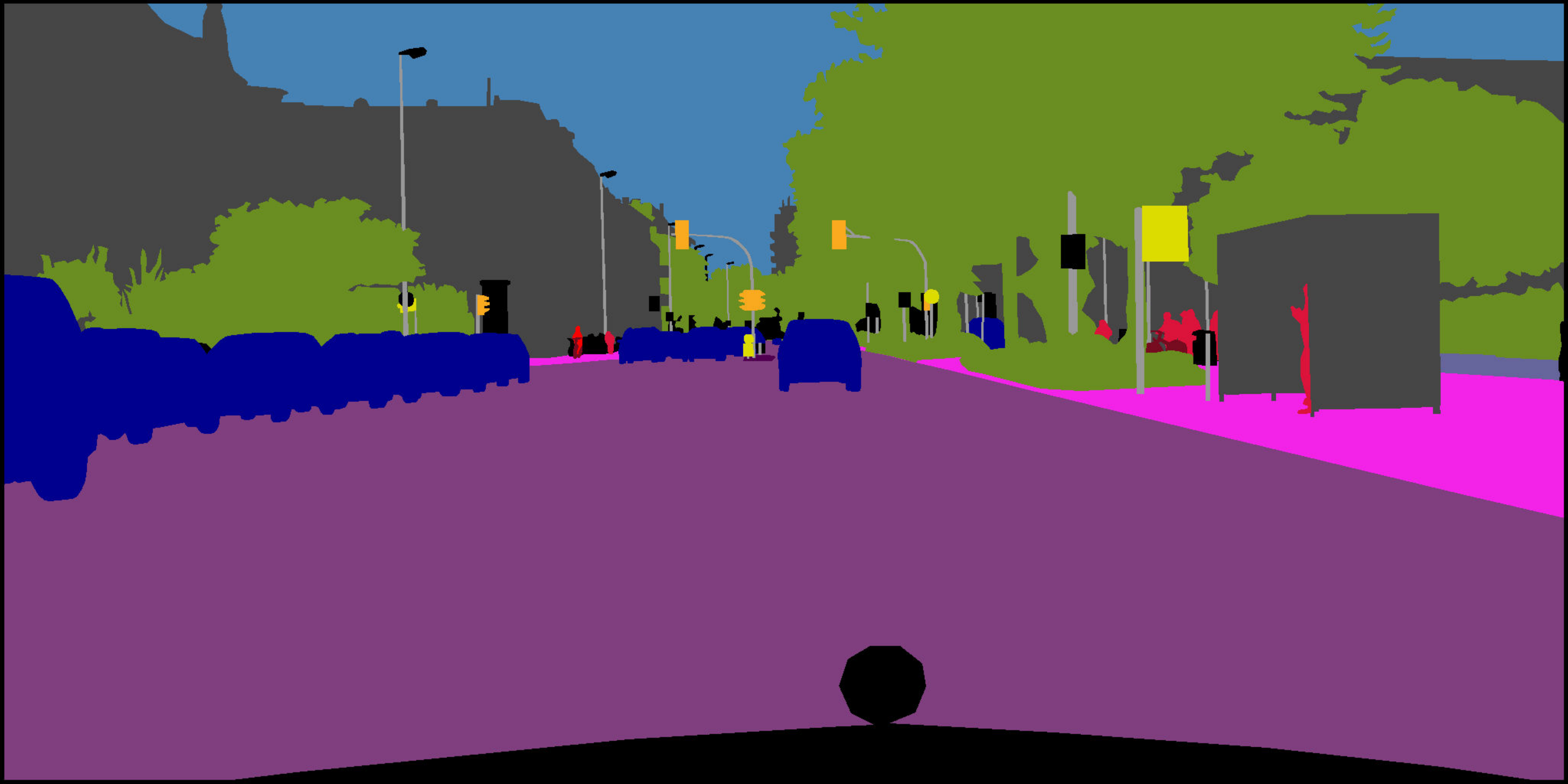}\\ \vspace{0.05cm}
		\includegraphics[width=0.49\linewidth]{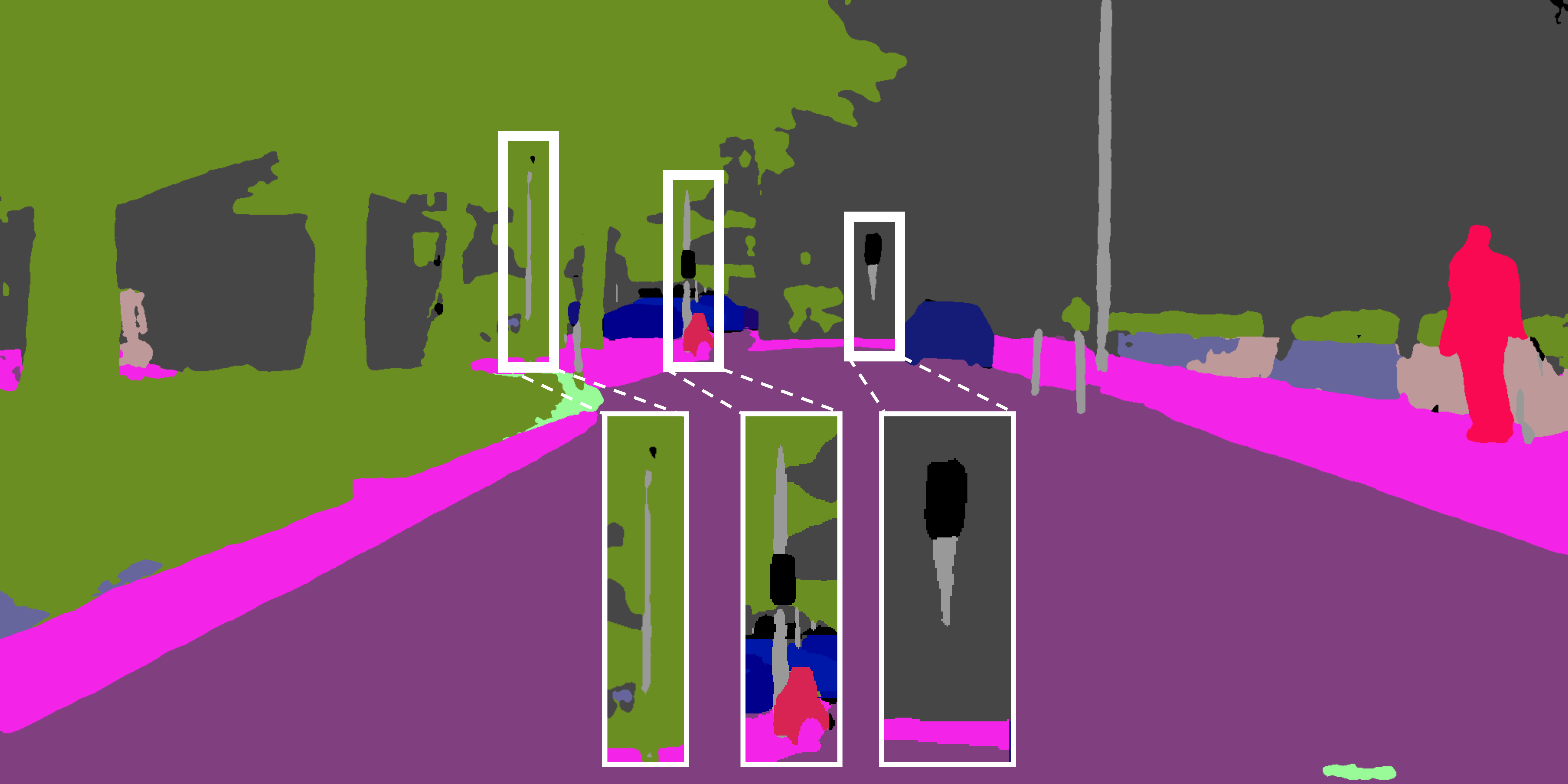}
		\includegraphics[width=0.49\linewidth]{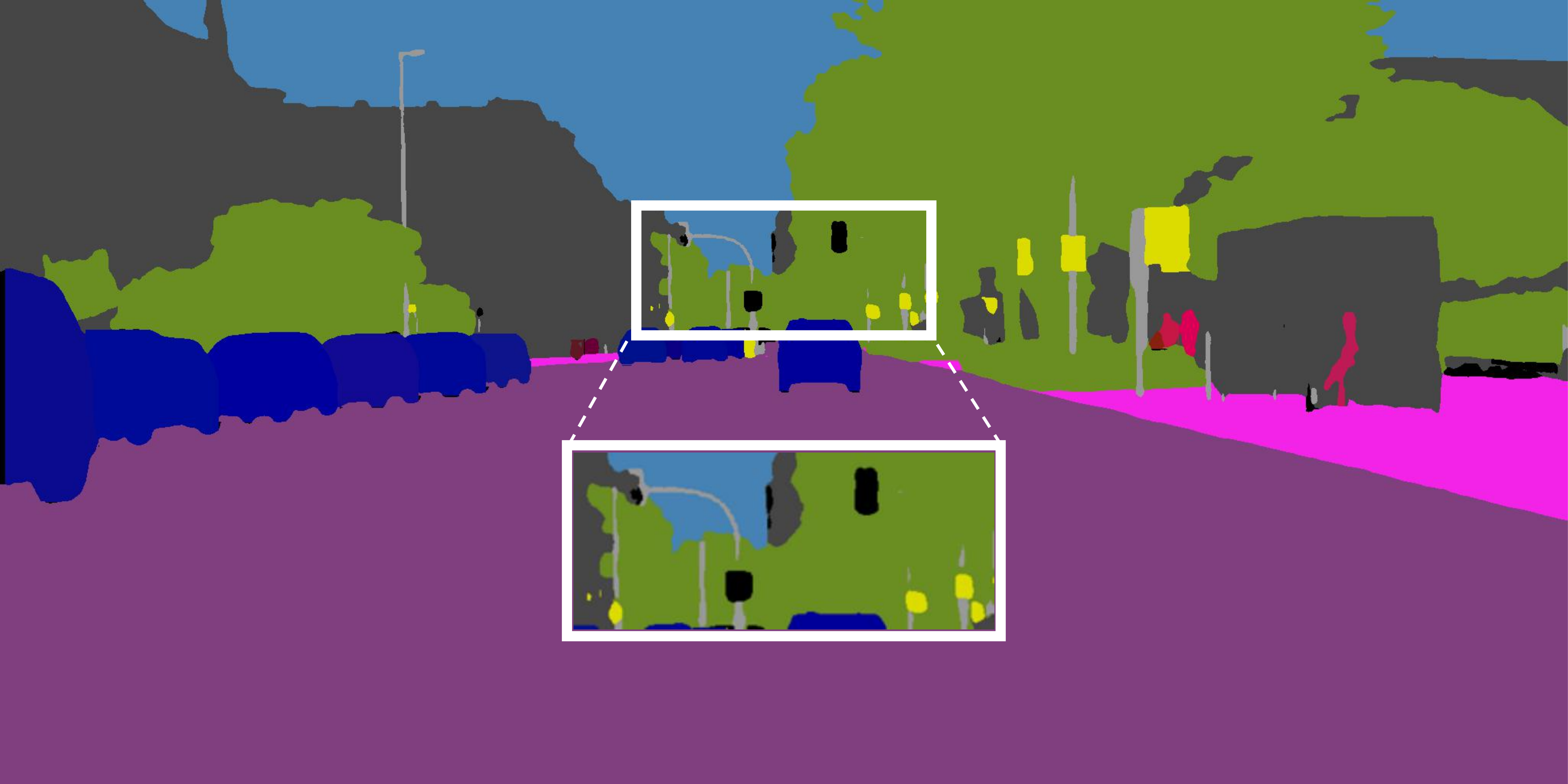}\\ \vspace{0.05cm}
		\includegraphics[width=0.49\linewidth]{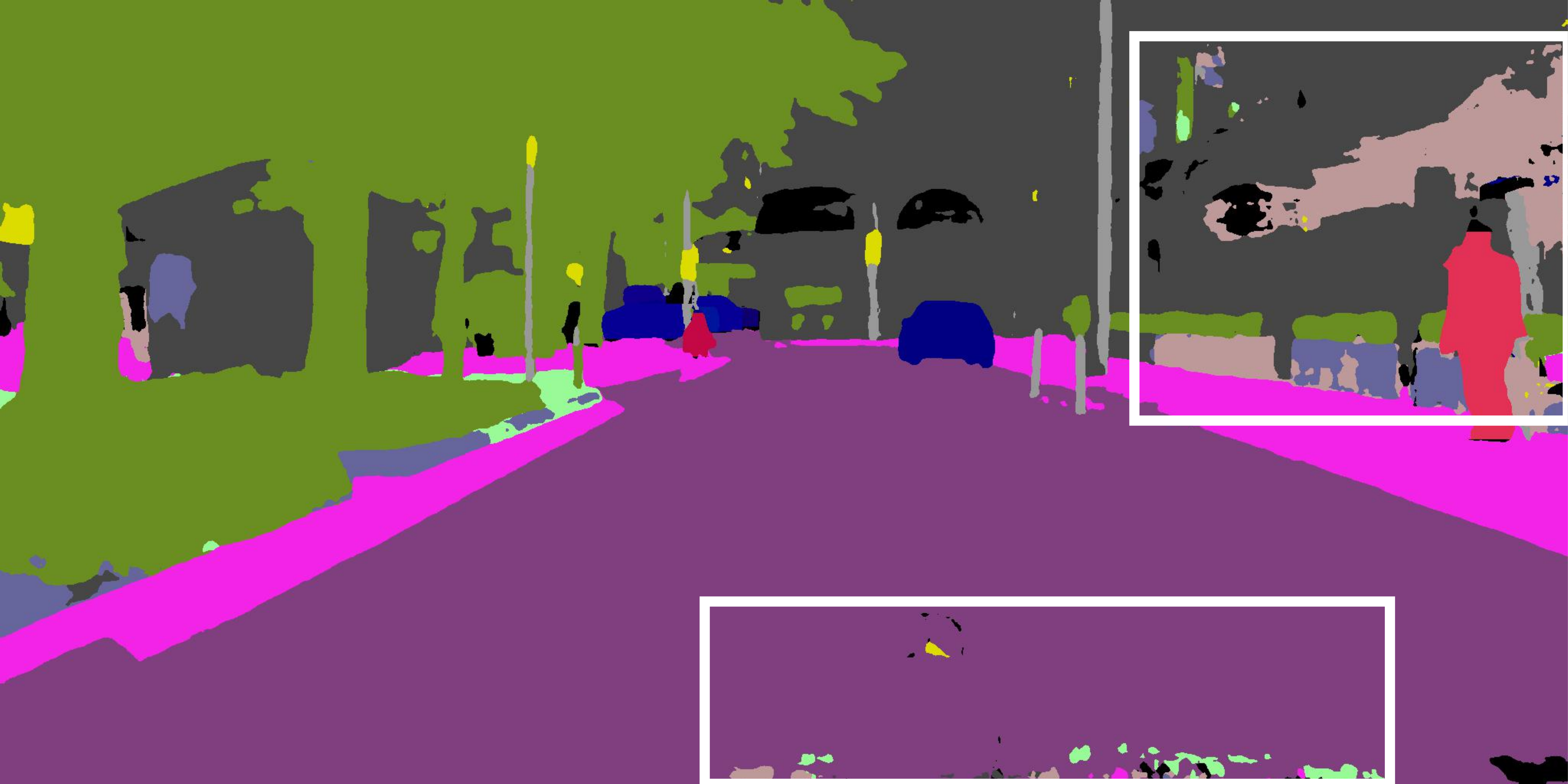}
		\includegraphics[width=0.49\linewidth]{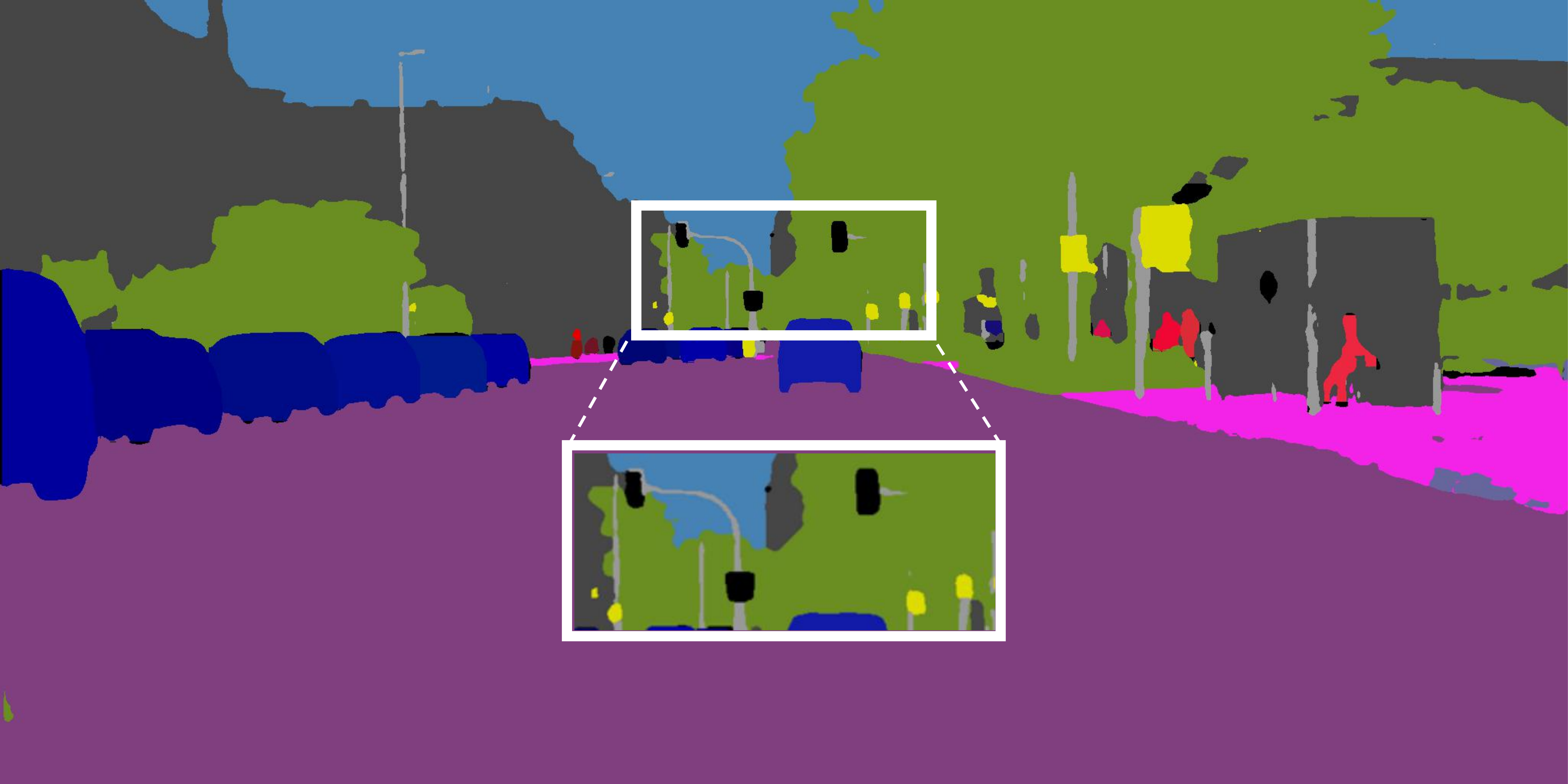}\\ \vspace{0.05cm}
		\includegraphics[width=0.49\linewidth]{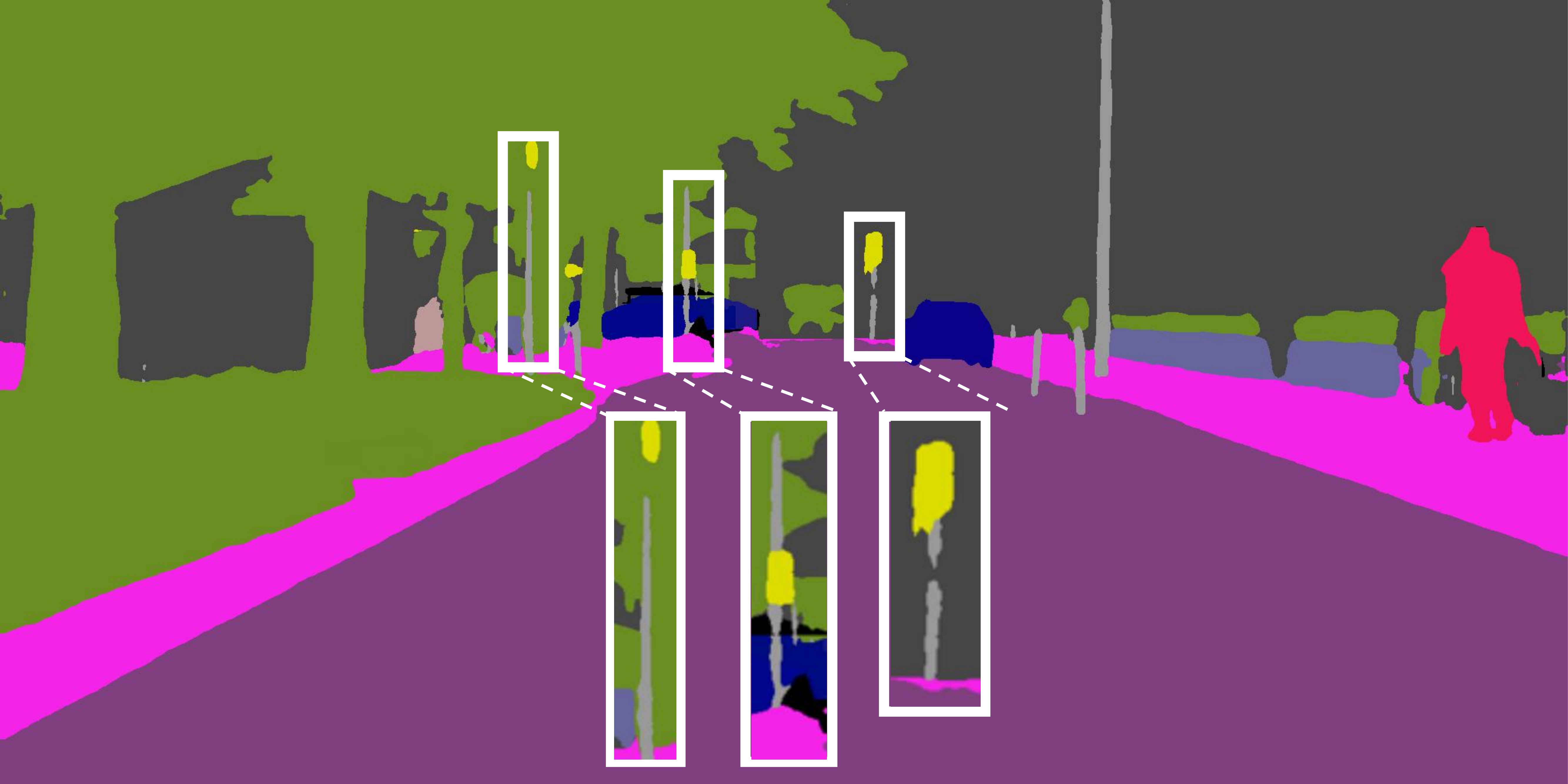}
		\includegraphics[width=0.49\linewidth]{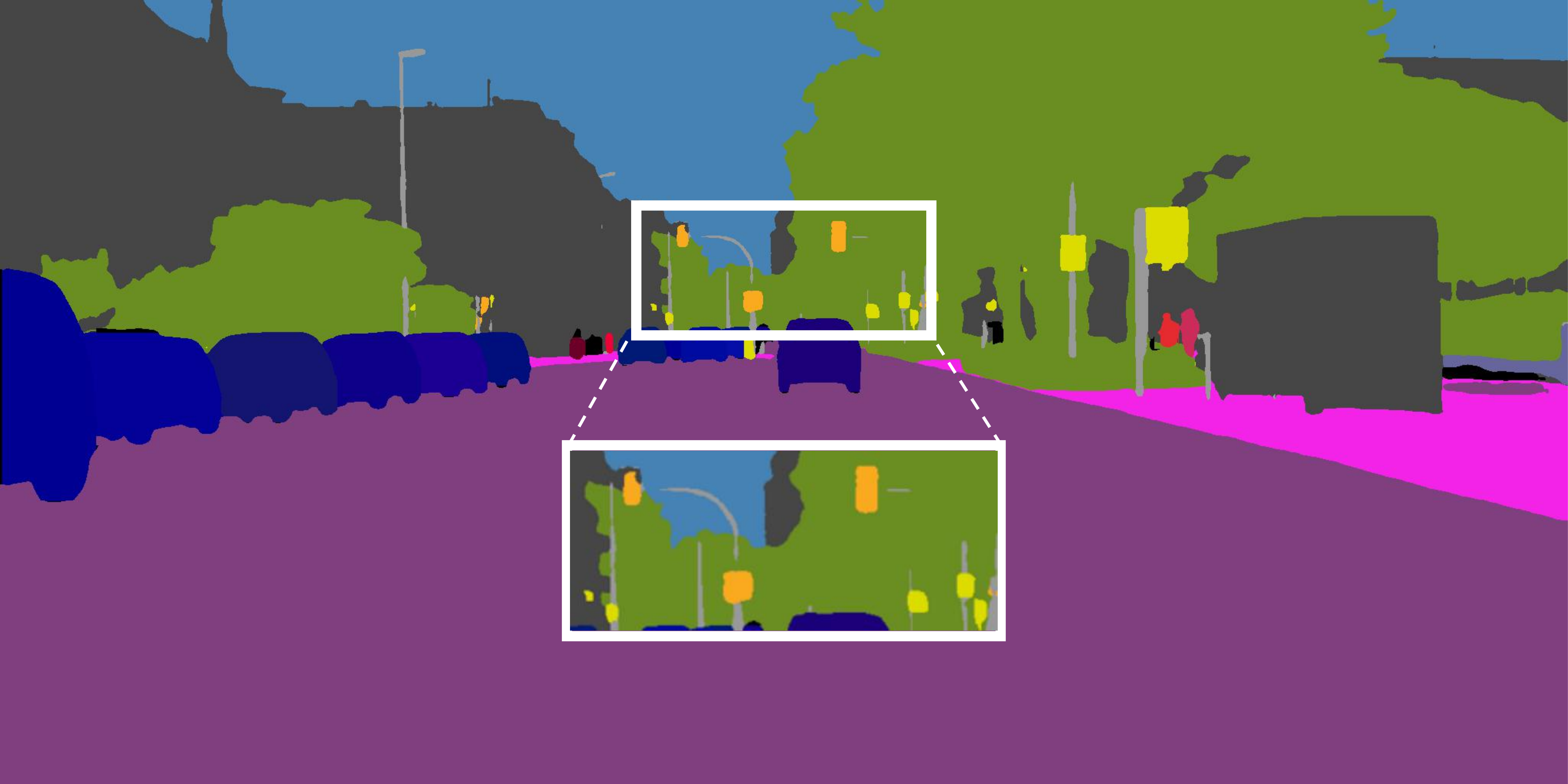}\\
	\caption{Qualitative segmentation results of panoptic segmentation models with different pyramid-like networks. 
	Problems are marked by white boxes, and small objects are scaled up to the road for better viewing. 
	From top to down are raw images, ground truth, predictions of UPSNet50-r, UPSNet50-r with 2-way FPN, UPSNet50-r with Convectional Network.
	Convectional Network helps model segment small-scale stuff.}
	\label{fig:Convectional Network_result}
\end{figure}

Fig.~\ref{fig:Convectional Network_result} shows qualitative panoptic segmentation results of panoptic segmentation models with different pyramid-like networks. 
From top to down are raw images, ground truth, predictions of UPSNet50-r,
UPSNet50-r with 2-way FPN, UPSNet50-r with Convectional Network, respectively.
From Fig.~\ref{fig:Convectional Network_result}, it can be seen that the model performs better on small-scale stuff like distant traffic signs and thin poles with Convectional Network. 

\begin{table*}[!ht]
	\renewcommand{\arraystretch}{1.0}
	\caption{Performance of Pixel-relation Block and Convectional Network on Panoptic FCN.
	\\The best value in each column is highlighted in bold}
	\label{tab:PB_FCN}
	\centering
	\begin{tabular}{c|c|c|c|c|c|c|c|c|c|c|c|c} \hline
	\multirow{2}{*}{method} & \multicolumn{3}{c|}{PQ} & \multicolumn{3}{c|}{SQ} & \multicolumn{3}{c|}{RQ} & AP & mIoU & \multirow{2}{*}{Params(M)}\\ \cline{2-12}
	 & all & thing & stuff & all & thing & stuff & all & thing & stuff & all & all & \\ \hline
	Panoptic FCN\cite{panopticfcn} & 57.6 & 50.3 & 63.0 & 79.5 & 77.7 & 80.8 & 71.2 & 64.5  & 76.1  & 29.5 & 75.8 & \bfseries{36.753}\\ \hline
	+Pixel-relation Block & \bfseries{59.8} & \bfseries{51.7} & \bfseries{65.7} & \bfseries{80.4} & \bfseries{78.8} & \bfseries{81.6} & \bfseries{73.3} & \bfseries{65.4} & \bfseries{79.0} & \bfseries{31.3} & 76.7 & 37.688 \\ \hline 
	+Convectional Network & 59.3 & 50.9 & 65.5 & \bfseries{80.4} & \bfseries{78.8} & 81.5 & 72.7 & 64.3 & 78.7 &31.0 & \bfseries{77.1} & 37.737 \\ \hline
\end{tabular}
\end{table*}

\begin{table*}[!ht]
	\renewcommand{\arraystretch}{1.0}
	\caption{Performance of SUNet on Cityscapes validation and test sets.\\The best value in each column is highlighted in bold}
	\label{tab:SUNet}
	\centering
	\begin{tabular}{c|c|c|c|c|c|c|c|c|c|c|c|c}
		\hline
		\multirow{2}{*}{Models} & \multirow{2}{*}{Backbone} & \multirow{2}{*}{Pretrain Dataset} & \multicolumn{5}{c|}{Validation Set} & \multicolumn{3}{c|}{Test Set} & \multirow{2}{*}{Params(M)} &\multirow{2}{*}{Time(ms)}\\ \cline{4-11}
		& & & PQ	& SQ	& RQ	&AP	&mIoU	& PQ	& SQ	& RQ	&	& \\ \hline
		UPSNet-r	& ResNet50	& ImageNet	& 58.1	& 79.5	& 71.7	& 32.6&75.0 & 55.2	& 79.2	& 68.2	& \bfseries{44.085}& \bfseries{275.6} \\ \hline
		\bfseries{SUNet} & ResNet50	& ImageNet	& \bfseries{61.7} & \bfseries{80.8}	& \bfseries{75.1} & \bfseries{35.3} & \bfseries{78.5} & \bfseries{58.1} & \bfseries{80.6} & \bfseries{70.9}	& 45.086  & 303.8 \\ \hline		\hline
		UPSNet-r	& ResNet101	& ImageNet+COCO	& 60.7	& 80.7	& 74.0	& 38.3 & 78.4 & 57.3 & 80.3	& 70.1 & \bfseries{63.077} & \bfseries{332.5}\\ \hline
		\bfseries{SUNet} & ResNet101 & ImageNet	& \bfseries{62.3} & \bfseries{81.1}	& \bfseries{75.6} & \bfseries{38.5} & \bfseries{79.5} & \bfseries{59.2}	& \bfseries{80.9} & \bfseries{72.0}	& 64.095 & 387.9 \\ \hline 
	\end{tabular}
	
\end{table*}

\begin{figure*}[ht]
	\centering
		\includegraphics[width=0.24\linewidth]{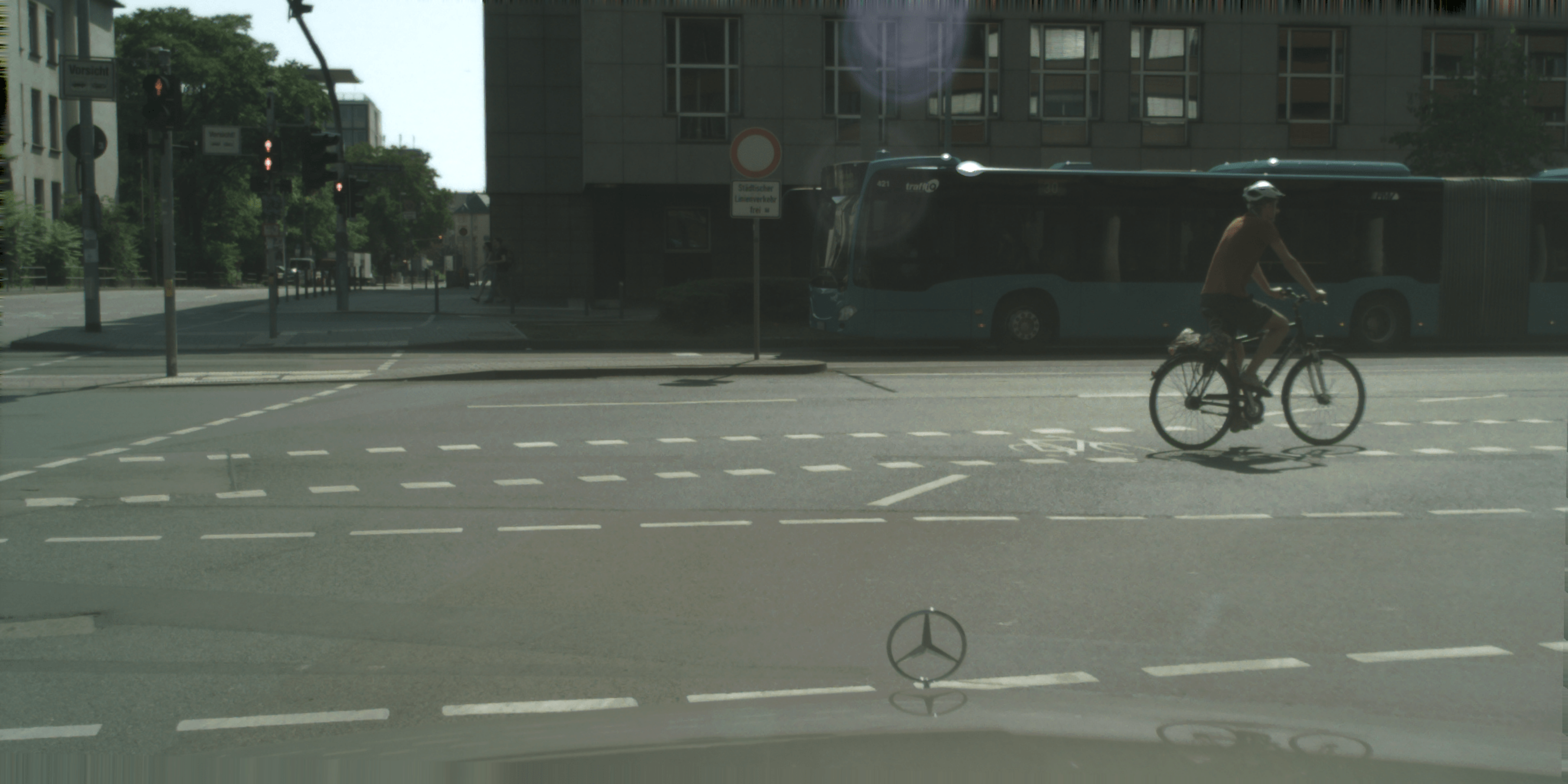}
		\includegraphics[width=0.24\linewidth]{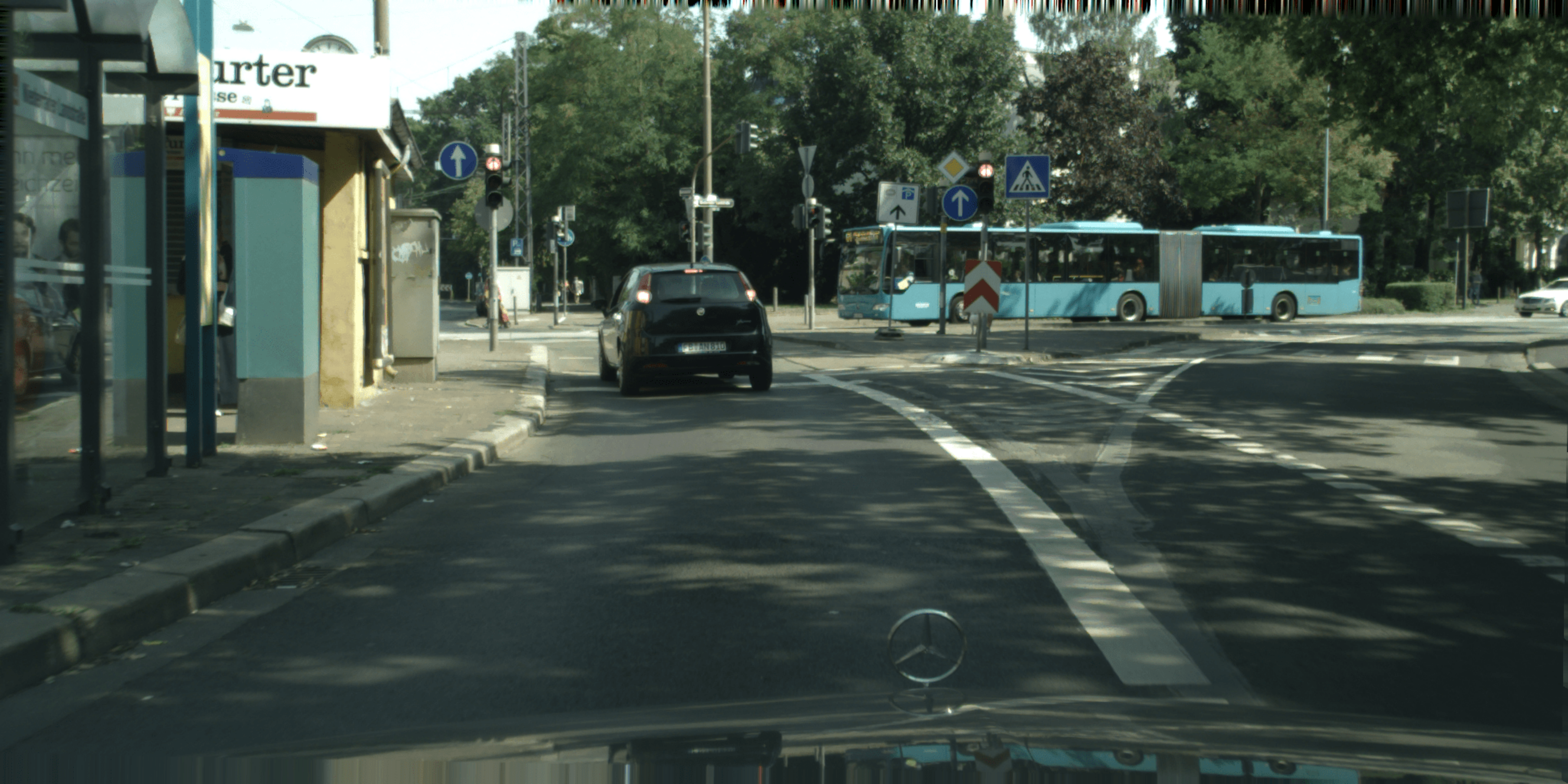}
		\includegraphics[width=0.24\linewidth]{figure/BiFPN/3/origin_3.pdf}
		\includegraphics[width=0.24\linewidth]{figure/BiFPN/13/origin_13.pdf}\\ \vspace{0.05cm}
		\includegraphics[width=0.24\linewidth]{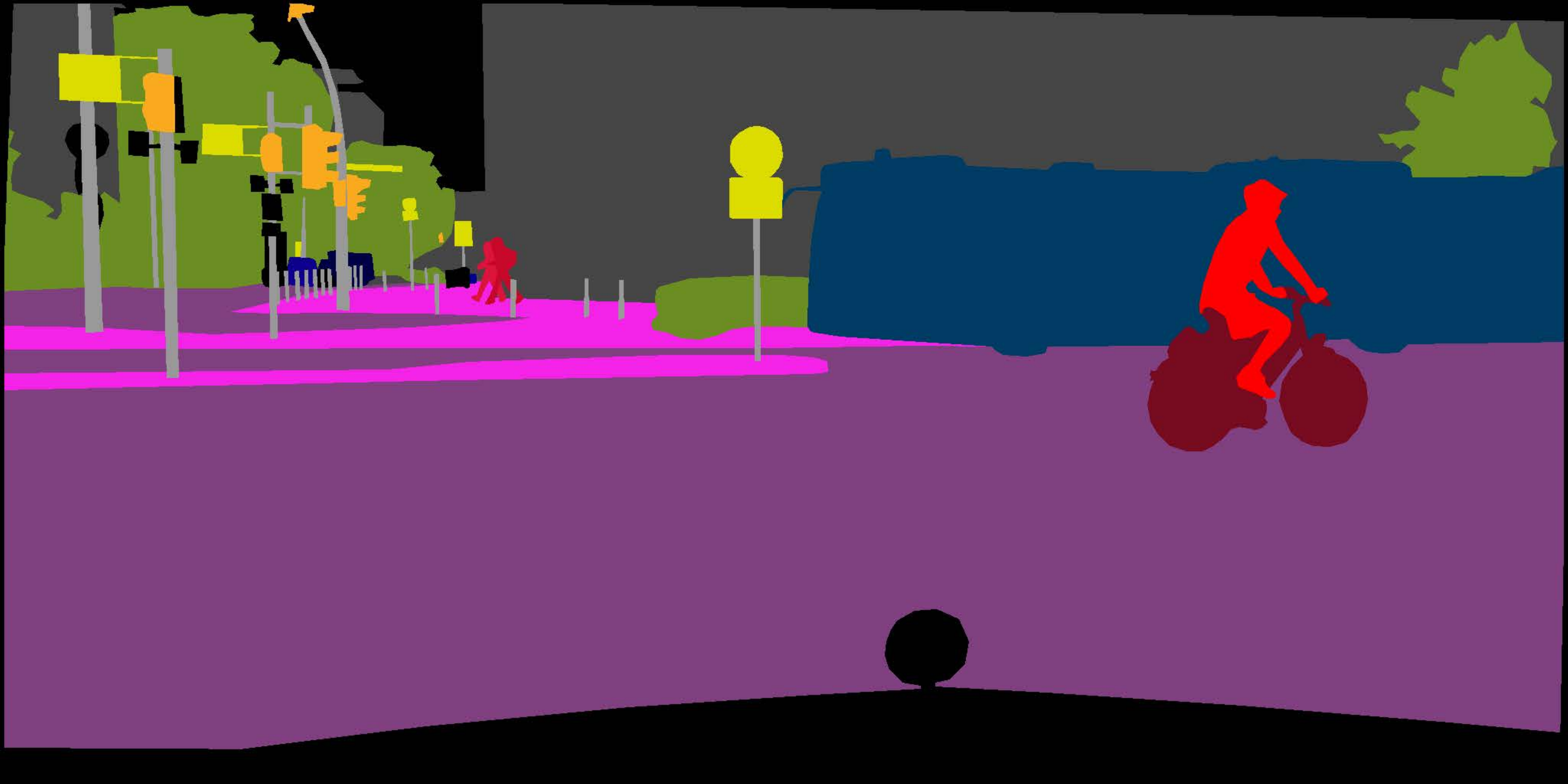}
		\includegraphics[width=0.24\linewidth]{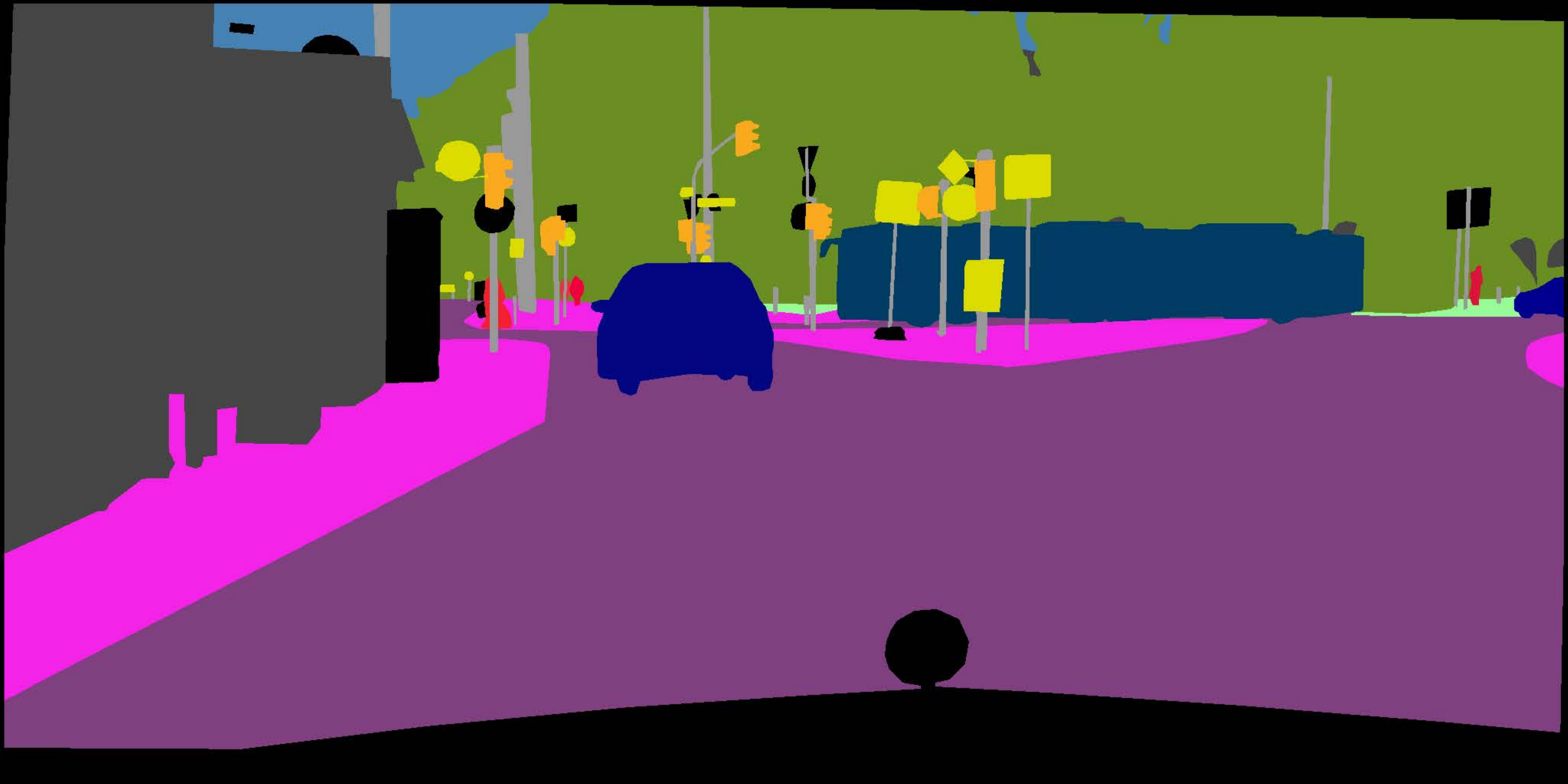}
		\includegraphics[width=0.24\linewidth]{figure/BiFPN/3/gtfine_3.pdf}
		\includegraphics[width=0.24\linewidth]{figure/BiFPN/13/gtfine_13.pdf}\\ \vspace{0.05cm}
		\includegraphics[width=0.24\linewidth]{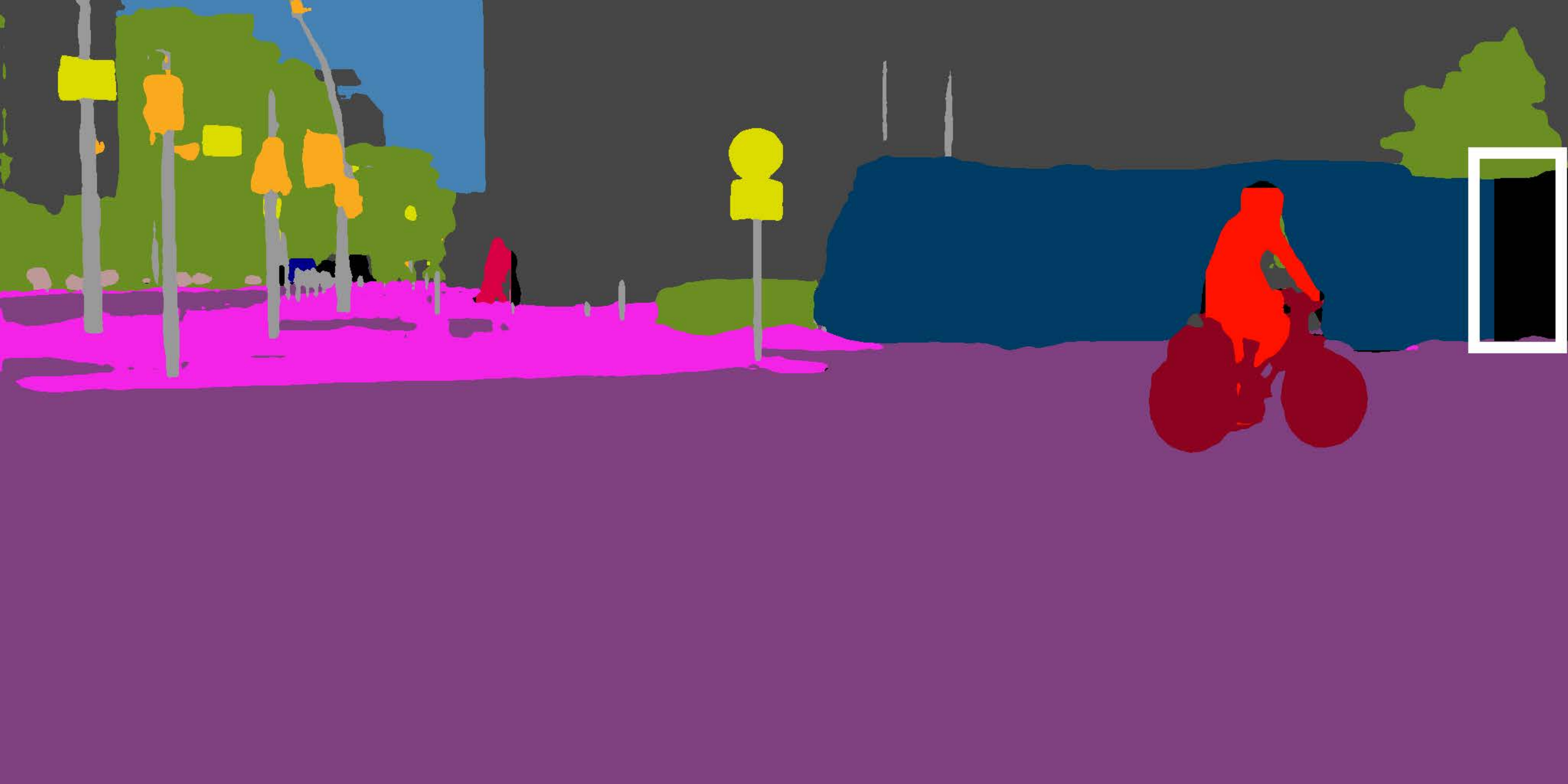}
		\includegraphics[width=0.24\linewidth]{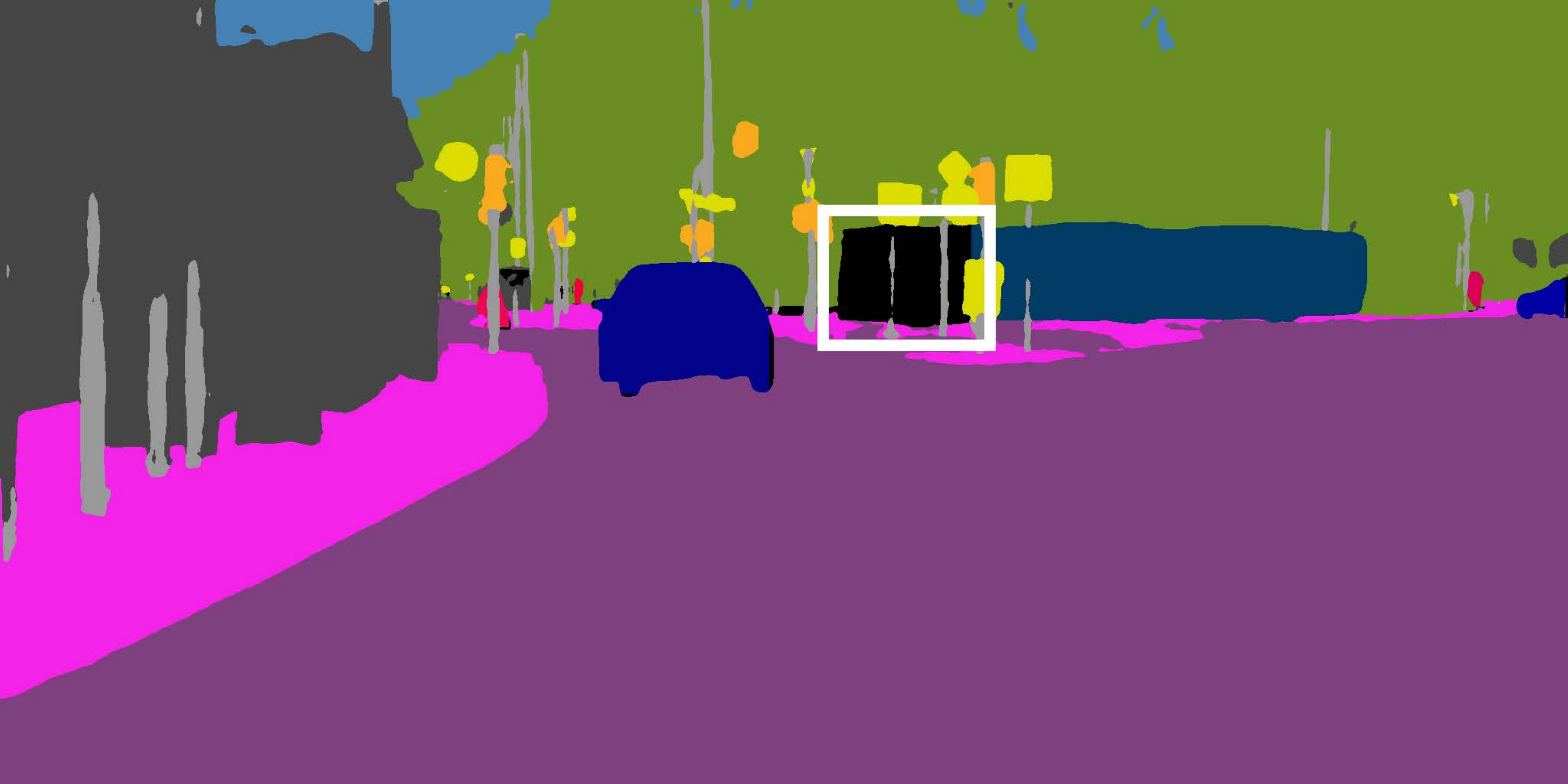}
		\includegraphics[width=0.24\linewidth]{figure/BiFPN/3/ups50_3.pdf}
		\includegraphics[width=0.24\linewidth]{figure/BiFPN/13/ups50_13.pdf}\\ \vspace{0.05cm}
		\includegraphics[width=0.24\linewidth]{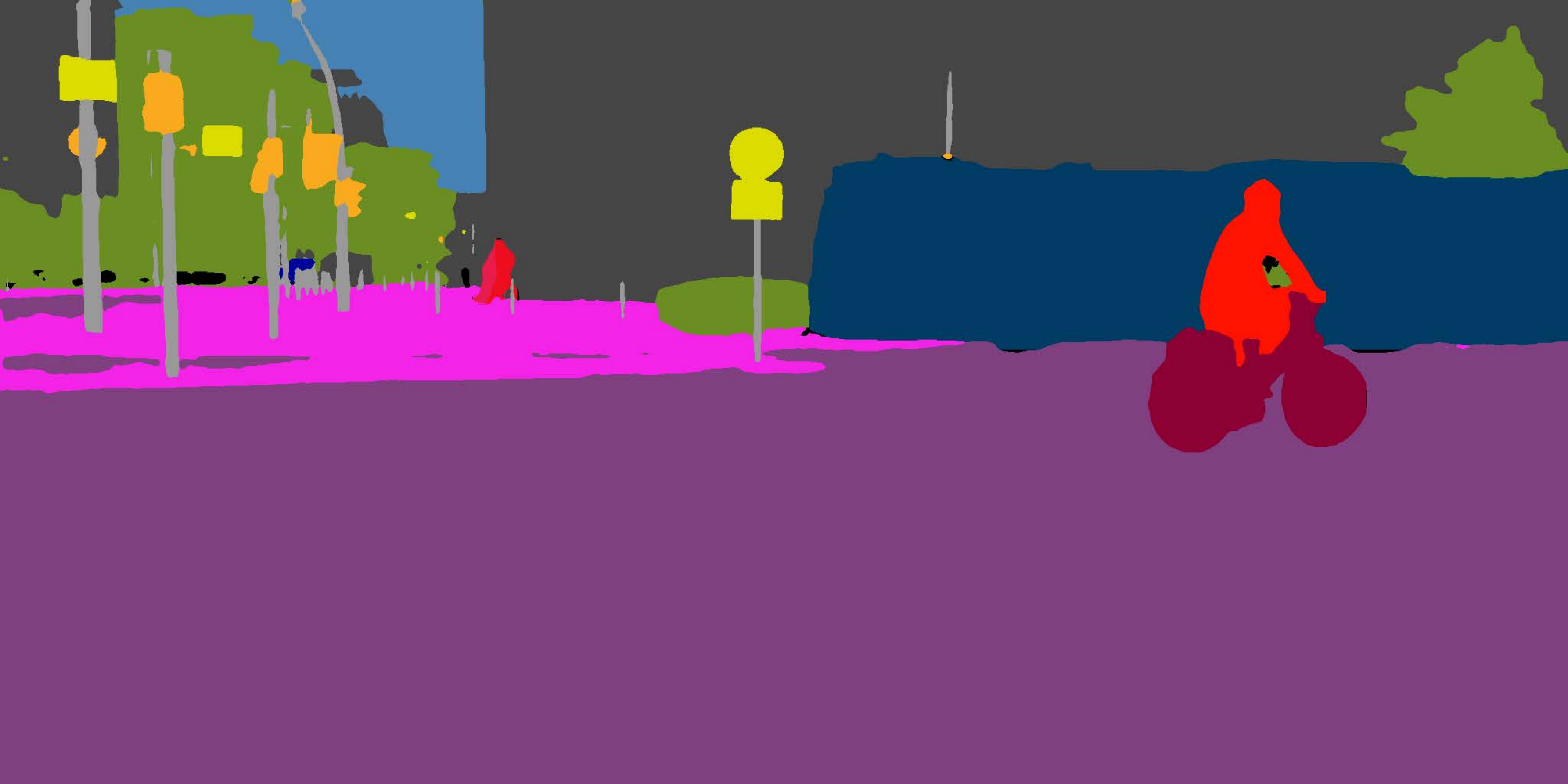}
		\includegraphics[width=0.24\linewidth]{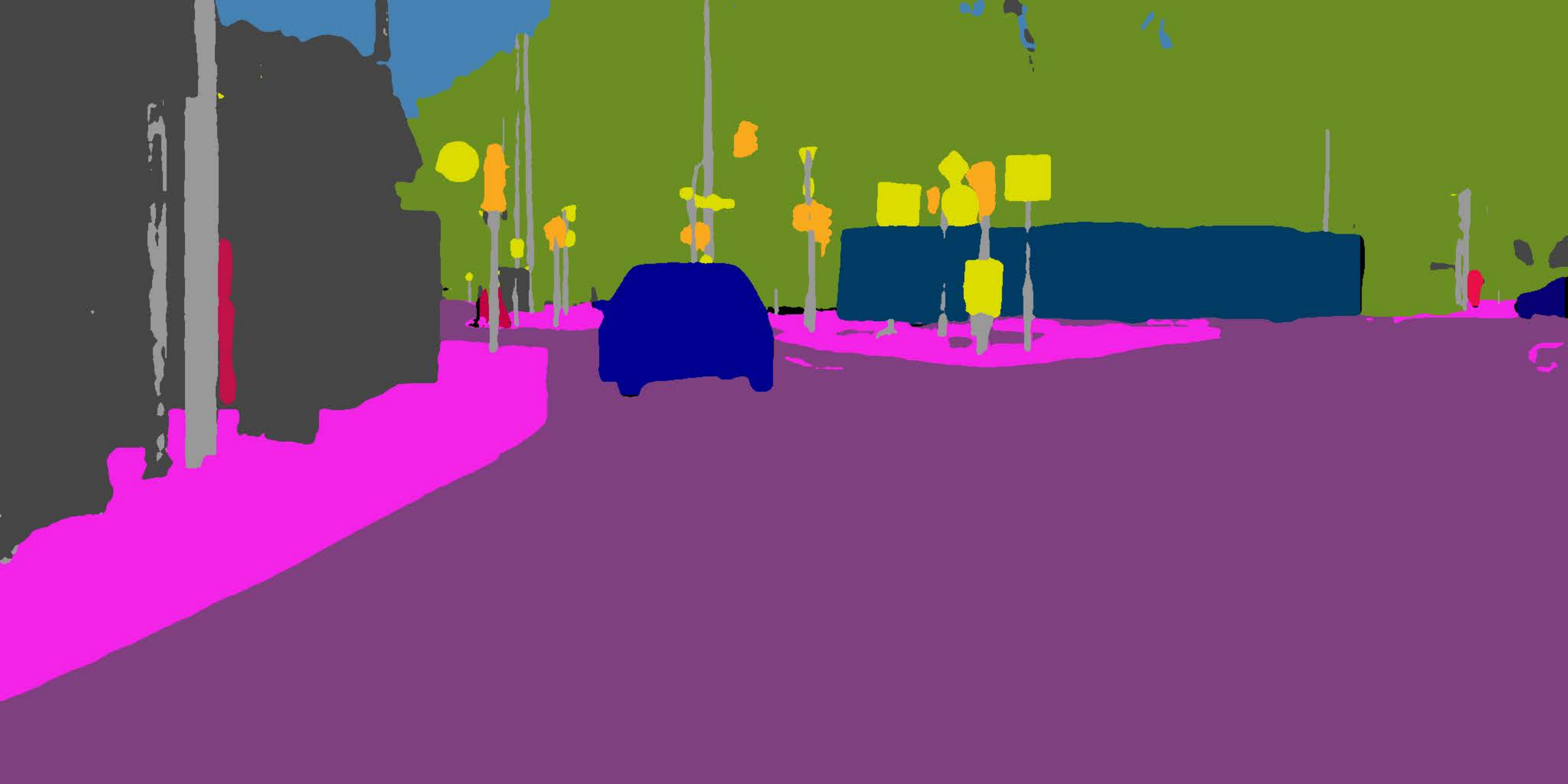}
		\includegraphics[width=0.24\linewidth]{figure/BiFPN/3/bifpn_3.pdf}
		\includegraphics[width=0.24\linewidth]{figure/BiFPN/13/bifpn_13.pdf}\\ 
	\caption{Qualitative panoptic segmentation results of SUNet50. Problems are marked by white boxes, and small objects are scaled up to the road for better viewing. 
	From top to down are raw images, ground truth, predictions of UPSNet50-r, and predictions of SUNet50, respectively.}
	\label{fig:SUNet_result}
\end{figure*}

Meanwhile, Convectional Network is also integrated into Panoptic FCN\cite{panopticfcn} like Pixel-relation Block to verify its generalization
and Table~\ref{tab:PB_FCN} presents the quantitative experimental results.
PQ, SQ, RQ, AP and mIoU of Panoptic FCN increase by 1.7, 0.9, 1.5, 1.5 and 1.3, respectively,
with only 0.984M extra parameters.
The steady improvement demonstrates the effectiveness of Convectional Network.


\subsection{Performance of SUNet and Comparison}

The performance of SUNet on the Cityscapes dataset is shown in Table~\ref{tab:SUNet}. 
For the validation set, SUNets exceed the UPSNets-r by 3.6 and 2.9 on PQ when using ResNet50 and ResNet101, respectively.
Meanwhile, it has higher SQ, RQ, AP, and mIoU on the validation set. 
The corresponding PQ improvements on the test set are 2.9 and 1.9.


The qualitative panoptic segmentation results of SUNet50 are visualized in Fig.~\ref{fig:SUNet_result}.
From top to down are raw images, ground truth, 
predictions of UPSNet50-r, and predictions of SUNet50, respectively.
White boxes mark problems, and small objects are scaled up to the road for better viewing.
It can be seen from the results that  
SUNet is more adaptable to multi-scale objects in the driving scene and performs better at extremely large and small ones.

For the prediction time, SUNet increases by 28.2 ms compared with UPSNet when ResNet50 is used.
When using ResNet101, the increased prediction time is 55.4 ms for more PB-ResNet Module in the backbone.

\begin{table*}[!ht]
	\renewcommand{\arraystretch}{1.0}
	\caption{Comparison of performance  on Cityscapes validation set.
	-m indicates that multiscale testing is used for inference.
	\\The highest and second highest values in each column are highlighted in bold}
	\label{tab:val_comparison}
	\centering
	\begin{tabular}{c|c|c|c|c|c|c|c|c}
		\hline
		\multirow{2}{*}{Models} &\multirow{2}{*}{Year} & \multirow{2}{*}{Backbone} & \multirow{2}{*}{Pretrain Dataset} & \multicolumn{3}{c|}{PQ} & {AP} & {mIoU}\\ \cline{5-9} 
		& & & & all & things & stuff & all & all \\ \hline
		Pixel Consus\cite{pixelvoting} &2020 & ResNet50 &ImageNet & 54.2 & 47.8 & 58.9 & - & 74.1\\ \hline
		FPSNet\cite{FastPSNet1} &2020 & ResNet50 &ImageNet & 55.1 & 48.3 & 60.1 & - & - \\ \hline
		AUNet\cite{li2019attention} & 2019 & ResNet50 &ImageNet & 56.4 & 52.7 & 59.0 & 33.6 & 73.6 \\ \hline 
		Axial-DeepLab\cite{axialdeeplab} & 2020 & Axial-Deeplab &ImageNet & 58.1 & - & - & 30.0 & 73.3 \\ \hline
		Real Time PS\cite{hou2020realtime} & 2020 & ResNet50 &ImageNet & 58.8 & 52.1 & 63.7 & 29.8 & 77.0 \\ \hline
		UPSNet\cite{utips} & 2019 & ResNet50 &ImageNet & 59.3 & 54.6 & 62.7 & 33.3 & 75.2 \\ \hline
		OCFusion\cite{lazarow2020learninginstance} &2020 & ResNet50 &ImageNet & 59.3 & 53.5 & 63.6 & - & - \\ \hline
		Panoptic FCN\cite{panopticfcn}&2021 & ResNet50 &ImageNet & 59.6 & 52.1 & 65.1 & - & - \\ \hline
		HLE\cite{2021hierarchical}&2021 & ResNet50 &ImageNet & 59.8 & 51.1 & 66.1 & - & - \\ \hline
		UPSNet\cite{xiong2019upsnet} &2019 &ResNet50 &ImageNet+COCO & 60.5 & \bfseries{57.0} & 63.0 & \bfseries{37.8} & 77.8 \\ \hline
		UTIPS\cite{utips} &2020 & ResNet50 &ImageNet & 61.4 & 54.7 & \bfseries{66.3} & {33.7} & \bfseries{79.5} \\ \hline
		COPS\cite{cops}&2021 & ResNet50 &ImageNet & \bfseries{62.1} & {55.1} & \bfseries{67.2} & - & -\\ \hline 
		\bfseries{SUNet(Ours)} & - & ResNet50 &ImageNet & \bfseries{61.7} & \bfseries{57.0} & 65.1 & \bfseries{35.3} & \bfseries{78.5}\\ \hline \hline
		Panoptic FPN\cite{panopticfpn}&2019 & ResNet101 &ImageNet & 58.1 & 52.0 & 62.5 & 33.0 & 75.7\\ \hline
		AUNet\cite{li2019attention} & 2019 & ResNet101 &ImageNet & 59.0 & 54.8 & 62.1 & 34.4 & 75.6 \\ \hline
		SpatialFlow\cite{spatialflow}&2020 & ResNet101 &ImageNet & 59.6 & 55.0 & 63.1 & - & - \\ \hline
		HLE\cite{2021hierarchical}&2021 & ResNet101 &ImageNet+COCO & 60.6 & 51.4 & \bfseries{67.2} & - & - \\ \hline
		UPSNet\cite{xiong2019upsnet} & 2019 & ResNet101 &ImageNet & 61.0 & {57.5} & 63.6 & \bfseries{39.0} & {77.9}\\ \hline
		Panoptic FCN\cite{panopticfcn}&2021 & ResNet101 &ImageNet & 61.4 & 54.8 & \bfseries{66.6} & - & - \\ \hline
		UPSNet-m \cite{xiong2019upsnet} &2019 & ResNet101 & ImageNet+COCO & 61.8 & 57.6 & 64.8 & \bfseries{39.0} & 79.2 \\ \hline
		AdaptIS-m\cite{adaptis} &2019 & ResNeXt101 &ImageNet & 62.0 & \bfseries{58.7} & 64.4 & 36.3 & {79.2} \\ \hline
		EfficientPS\cite{efficientps}&2021 &EfficientNet &ImageNet & \bfseries{63.9} & \bfseries{60.7} & 66.2 & {38.3} & \bfseries{79.3} \\ \hline
		\bfseries{SUNet(Ours)}&- & ResNet101 &ImageNet & \bfseries{62.3} & 57.4 & 65.8 & {38.5} & \bfseries{79.5} \\ \hline
	\end{tabular} 
\end{table*}

The comparison results of the models' performance on Cityscapes and COCO are shown in Table~\ref{tab:val_comparison} and Table~\ref{tab:coco_val_comparison}, respectively. 
For convenience, R50 and AD represent ResNet50 and Axial-DeepLab\cite{axialdeeplab} in Table~\ref{tab:coco_val_comparison}. 
It should be noted that many methods only show the performance on the validation set,
and we mainly compare the methods that adopt ResNet as the backbone for a fair comparison. 

\begin{table}[!t]
	\renewcommand{\arraystretch}{1.0}
	\caption{Comparison of performance  on COCO validation set.
	\\R50 and AD represent ResNet50 and Axial-DeepLab.
	\\The highest and second highest values in each column are highlighted in bold}
	\label{tab:coco_val_comparison}
	\centering
	\resizebox{\linewidth}{2.21cm}{
	\begin{tabular}{c|c|c|c|c|c|c|c}
		\hline
		\multirow{2}{*}{Models}& \multirow{2}{*}{Year} & \multirow{2}{*}{Backbone} & \multicolumn{3}{c|}{PQ} & {SQ} & {RQ}\\ \cline{4-8} 
		& & & all & things & stuff & all & all \\ \hline
		PCV\cite{pixelvoting} &2020 &R50 &37.5 &40.0 &33.7 & 77.7 &47.2 \\ \hline
		COPS\cite{cops} &2021 & R50 & 38.4 & 40.5 & 35.2 & - & - \\ \hline
		Panoptic FPN\cite{panopticfpn} & 2019 & R50 & 39.4 & 45.9 & 29.6 & 77.8 & 48.3 \\ \hline
		CIAE\cite{ciae} &2021 & R50 & 40.2 & 45.3 & 32.3 & - & - \\ \hline
		UPSNet\cite{xiong2019upsnet} &2019 & R50 & 42.5 & 48.6 & 33.4 & 78.0 & 52.5 \\ \hline
		CQB-Net\cite{du2021save} &2021 &R50 & 42.7 & 49.5 & 32.3 & - & - \\ \hline
		UTIPS\cite{utips} &2020 & R50 & 43.4 & 48.6 & 35.5 & 79.6 & 53.0 \\ \hline
		Axial-Deeplab\cite{axialdeeplab} &2020 & AD & 43.9 & 48.6 & \bfseries{36.8} & - & - \\ \hline
		Panoptic FCN\cite{panopticfcn} &2021 & R50 & 44.3 & 50.0 & 35.6 & \bfseries{80.7} & 53.0 \\ \hline
		MaskFormer\cite{maskformer} &2021 & R50+6Enc & \bfseries{46.5} & \bfseries{51.0} & \bfseries{39.8} & \bfseries{80.4} & \bfseries{56.8} \\ \hline
		\bfseries{SUNet(Ours)} &- & R50 & \bfseries{45.0} & \bfseries{51.3} & {35.7} & \bfseries{80.4} & \bfseries{54.9} \\ \hline
	\end{tabular} 
 	}
\end{table}

Though without complicated backbones and large-scale datasets for pretraining, 
SUNet50 and SUNet101 achieve 61.7 and 62.3 of PQ on the Cityscapes validation set and have competitive AP and mIoU values. 
Moreover, SUNet50 obtains a PQ of 45.0 on the COCO validation set. 
It outperforms most existing methods adopting ResNet50 as the backbone except for MaskFormer\cite{maskformer}, 
which uses additional six encoders and transformer architecture.
The experimental results generally demonstrate that SUNet has competitive panoptic segmentation performance on Cityscapes and COCO datasets. 


\subsection{Ablation Study}
\label{ablation}
In \ref{LSB} and \ref{CN}, we have conducted ablation experiments on the Pixel-relation Block and Convectional Network. 
Here we explore the impact of three coefficients in the loss function on the SUNet50's performance on the Cityscapes validation set.
We set $\lambda_{i}=1.0$ and adjust the $\lambda_{s}$ and $\lambda_{p}$.
The experimental results are shown in Table \ref{tab:ablation loss}.

First, we fix $\lambda_{s}$ to 1.0, then set the $\lambda_{p}$ to 0.3, 0.5, 0.8, and 1.0, respectively.
Among them, $\lambda_{p}=0.5$ performs best and is selected, achieving a PQ value of 60.9.
After that, $\lambda_{s}$ is gradually increased to 1.2, 1.5, and 2.0 in the experiments, with 1.5 performing the best.
Finally, $\lambda_{s}$, $\lambda_{i}$, and $\lambda_{p}$ are tuned as 1.5, 1.0, and 0.5,
which brings a PQ value improvement of 0.9 compared with the settings of 1.0, 1.0, and 1.0.

\begin{table}[ht]
	\renewcommand{\arraystretch}{1.1}
	\caption{Ablation Study of Coefficients of Loss Functions. 
	\\$\lambda_{s}$, $\lambda_{i}$ and $\lambda_{p}$ correspond to Semantic Stuff Branch, Instance Thing Branch and Panoptic Fusion Module, respectively.
	\\The best values in each column are highlighted in bold}
	\label{tab:ablation loss}
	\centering
	\resizebox{\linewidth}{1.62cm}{	
		\begin{tabular}{ccc|c|c|c|c|c|c|c}
			\hline
			\multirow{2}{*}{$\lambda_{s}$} & \multirow{2}{*}{$\lambda_{i}$} & \multirow{2}{*}{$\lambda_{p}$} & \multicolumn{3}{c|}{PQ} & SQ & RQ & AP & mIoU \\ \cline{4-10}
			 & & &all &things &stuff & all& all& all& all \\ \hline 
			1.0 & 1.0 & 0.3 & 60.3 & 54.5 & 64.5 & \bfseries{80.9} & 73.3 & 34.2 & 78.4 \\ \hline
			1.0 & 1.0 & 0.5 & 60.9 & 55.3 & 64.9 & 80.7 & 74.2 & 34.6 & 78.3 \\ \hline
			1.0 & 1.0 & 0.8 & 60.6 & 55.3 & 64.4 & 80.6 & 73.9 & 34.0 & 77.5 \\ \hline
			1.0 & 1.0 & 1.0 & 60.8 & 55.4 & 64.8 & 80.7 & 74.1 & 34.3 & \bfseries{78.5} \\ \hline
			1.2 & 1.0 & 0.5 & 61.1 & 56.1 & 64.7 & 80.7 & 74.5 & 34.4 & 78.1 \\ \hline
			1.5 & 1.0 & 0.5 & \bfseries{61.7} & \bfseries{57.0} & \bfseries{65.1} & 80.8 & \bfseries{75.1} & \bfseries{35.3} & \bfseries{78.5} \\ \hline
			2.0 & 1.0 & 0.5 & 60.4 & 54.2 & 65.0 & 80.6 & 73.8 & 34.3 & 77.3 \\ \hline
		\end{tabular}}
\end{table}

\subsection{Discussion}
Both Pixel-relation Block and Convectional Network achieve the expected improvement effect through experiments.
Pixel-relation Block makes panoptic segmentation model predict less truncated large-scale things,
while Convectional Network improves segmentation performance on small-scale stuff like thin poles and distant traffic signs. 
Meanwhile, these two modules can conveniently be appended to other panoptic segmentation models by modifying the backbone.
Moreover, Pixel-relation Block can be used in instance branch or top layers. 
In our experiment, they only have been validated to be effective on Panoptic FCN\cite{panopticfcn} due to the limitation of computing resources.
Our future work will be to explore the performance and generalization of Pixel-relation Block and Convectional Network on more panoptic segmentation models.

Though SUNet has a competitive performance on Cityscapes dataset,
it still has a gap of about 3 points on PQ between validation and test set of Cityscapes. 
It indicates that the generalization of SUNet needs to be improved. 
Meanwhile, the pretraining model on Mapillary is a preferable method for better performance on Cityscapes\cite{panopticdeeplab,efficientps}.
For computation limitations, this method has not been adopted in our experiments, 
and we only tune the coefficients of loss function when training SUNet50 on Cityscapes.
In the future, we will consider conducting experiments on the Mapillary dataset to further verify the effectiveness of our methods.

For the COCO dataset, we only use ResNet50 as the backbone network for experiments 
due to a large amount of data and limited computing resources, 
requiring about a week of training on two 1080Ti GPUs.
Even without fine-tuning the loss function parameters, 
SUNet50 still achieves a competitive PQ value of 45.0 on the COCO validation set, showing the effectiveness of our method.

\section{Conclusion}
In this paper, we propose two lightweight modules to improve the performance of panoptic segmentation models for multi-scale objects.
Pixel-relation Block introduces global context information into panoptic segmentation models, 
which mitigates the problem of truncated detection of large-scale objects. 
Convectional Network is designed to supply high-resolution information for segmentation on small-scale stuff
and is appropriate for sharing by the downstream segmentation branches. 
Both of them achieve the expected improvement and can be conveniently utilized by panoptic segmentation models. 

Moreover, SUNet for panoptic segmentation based on Pixel-relation Block and Convectional Network is constructed. 
Experimental results demonstrate that SUNet performs better on multi-scale objects 
and achieves competitive panoptic segmentation performance on Cityscapes and COCO datasets.

\bibliographystyle{IEEEtran}
\bibliography{IEEEabrv,reference.bib}

\end{document}